%% file: paper.tex
\documentclass{article}

\usepackage{microtype}
\usepackage{graphicx}
\usepackage{subcaption}
\usepackage{booktabs} 

\usepackage{hyperref}

\usepackage[preprint]{icml2026}

\usepackage{amsmath}
\usepackage{amssymb}
\usepackage{mathtools}
\usepackage{amsthm}

\usepackage[capitalize,noabbrev]{cleveref}

\input{acronyms}

\input{macros}

\theoremstyle{plain}
\newtheorem{theorem}{Theorem}[section]

\theoremstyle{definition}

\newtheorem{assumption}[theorem]{Assumption}
\theoremstyle{remark}
\newtheorem{remark}[theorem]{Remark}

\usepackage[textsize=tiny]{todonotes}

\icmltitlerunning{Online Bayesian Experimental Design for Partially Observable Dynamical Systems}

\begin{document}

\twocolumn[
  \icmltitle{Online Bayesian Experimental Design for Partially Observed Dynamical Systems}



  \icmlsetsymbol{equal}{*}

  \begin{icmlauthorlist}
    \icmlauthor{Sara Pérez-Vieites}{aalto}
    \icmlauthor{Sahel Iqbal}{aalto}
    \icmlauthor{Simo Särkkä}{aalto}
    \icmlauthor{Dominik Baumann}{aalto}
  \end{icmlauthorlist}

  \icmlaffiliation{aalto}{Department of Electrical Engineering and Automation, Aalto University, Espoo, Finland}

  \icmlcorrespondingauthor{Sara Pérez-Vieites}{sara.perezvieites@aalto.fi}

  \icmlkeywords{Bayesian experimental design, Bayesian inference, Monte Carlo}

  \vskip 0.3in
]



\printAffiliationsAndNotice{}  

\begin{abstract}
  Bayesian experimental design (BED) provides a principled framework for optimizing data collection by choosing experiments that are maximally informative about unknown parameters. However, existing methods cannot deal with the joint challenge of (a) \emph{partially observable dynamical systems}, where only noisy and incomplete observations are available, and (b) \emph{fully online inference}, which updates posterior distributions and selects designs sequentially in a computationally efficient manner.
  Under partial observability, dynamical systems are naturally modeled as state-space models (SSMs), where latent states mediate the link between parameters and data, making the likelihood---and thus information-theoretic objectives like the expected information gain (EIG)---intractable. 
  We address these challenges by deriving new estimators of the EIG and its gradient that explicitly marginalize latent states, enabling scalable stochastic optimization in nonlinear SSMs. Our approach leverages nested particle filters for efficient online state-parameter inference with convergence guarantees. 
  Applications to realistic models, such as the susceptible–infectious–recovered (SIR) and a moving source location task, show that our method successfully handles both partial observability and online inference.
\end{abstract}

\section{Introduction}
\label{introduction}

Designing experiments that yield maximally informative data is a fundamental problem across science and engineering. The ability to guide data collection efficiently can accelerate learning in diverse domains, including material discovery \citep{lei2021bayesian,lookman2019active}, DNA sequencing \citep{weilguny2023dynamic}, sensor networks \citep{wu2023offline}, and drug discovery \citep{lyu2019ultra}. Formally, given a known model structure, let $\btheta$ represent the parameters of interest and $\bxi$ the design chosen for the experiment, which determines the corresponding observation $\by$. The objective is to select designs that maximize the information gained about $\btheta$, thereby reducing uncertainty as more data becomes available.

\Gls*{bed} provides a principled framework for combining Bayesian inference with data acquisition \citep{lindley1956measure,huan2024optimal}. In this setting, prior beliefs about the parameters $p(\btheta)$ are updated as data are collected, yielding a posterior distribution that quantifies uncertainty. This update is driven by the likelihood $p(\by\!\mid\! \btheta,\bxi)$, which links parameters to observations under a given design $\bxi$. 
When this process is repeated adaptively, using past observations to guide future design choices, we obtain the sequential \gls*{bed} setting 
\citep{Rainforth24}. At the experiment $t$, the prior is $p(\btheta \!\mid\! h_{t-1})$, where $h_{t-1}=\{ \bxi_{1:t-1}, \by_{1:t-1}\}$ is the history of past designs and observations. The next design $\bxi_t$
is then chosen to maximize some utility function, commonly the \gls*{eig}, which requires evaluating $p(\by_t \!\mid\! \btheta,\bxi_t)$.

A common assumption in sequential \gls*{bed} is that the likelihood $p(\by_t \!\mid\! \btheta,\bxi_t)$ is tractable and available in closed form. While this holds in some settings, it is unrealistic in many real-world scenarios where the system cannot be fully observed. In particular, dynamical systems often involve latent states that evolve over time and are only observed through noisy, partial measurements. Such systems are naturally modeled as \glspl*{ssm}~\citep{Sarkka23}, where the likelihood is defined only indirectly through integration over the hidden states, and is therefore \emph{intractable}.

Several \gls*{bed} approaches address \emph{implicit} likelihoods, i.e., cases where we can simulate observations from the model but cannot evaluate $p(\by_t \!\mid\! \btheta, \bxi_t)$ in closed form. In principle, such methods could also be applied or extended to \glspl*{ssm}, since they provide a way to simulate samples. 
Most existing implicit \gls*{bed} approaches rely on \emph{amortization}, where a surrogate (e.g., a neural network) is trained offline on a fixed dataset to approximate the likelihood, and then reused at deployment. Such methods have been proposed, mostly in static \gls*{bed} settings ~\citep{Kleinegesse20,Dehideniya18}, with recent extensions to sequential settings via mutual-information bounds~\citep{Ivanova21} or density-ratio estimation~\citep{Kleinegesse21}.
%


In contrast, we focus on a \emph{non-amortized, fully online} regime: designs are optimized sequentially using the current posterior and without offline training. This regime is motivated by applications where (i) decisions are made over long horizons, (ii) the system is only partially observed, and (iii) computational budget between measurements can vary---often prioritizing reliability over ultra-fast deployment, e.g., adaptive clinical trials \citep{chaloner1995bayesian} and astronomical observations \citep{loredo2004bayesian}.
A key practical issue in such settings is \emph{long-horizon feasibility}. In partially observed dynamical models, robust inference/design schemes that repeatedly rely on the full data history and long latent trajectories can incur computation and memory costs that grow with $T$, which may become prohibitive in practice. Our aim is therefore to develop methods that keep inference and design updates \emph{online} (with linear cost in $T$ and constant memory in the horizon length) while retaining convergence guarantees for the estimators used in design optimization. We discuss these trade-offs and connections to prior work in Section~\ref{sec:related work}.

\textbf{Contributions.}
In this paper, we propose a novel method for sequential \gls*{bed} in partially observable dynamical systems, where likelihoods are intractable and online inference is required. Our main contributions are 
(i) \emph{new estimators} of \gls*{eig} and its gradient, specifically derived for state-space models with latent states, enabling design optimization under partial observability; 
(ii) an \emph{online} sequential \gls*{bed} approach that couples these estimators with nested particle filters~\citep[NPFs,][]{Crisan18bernoulli}, which recursively updates the joint parameter--state posterior without reprocessing the full history, yielding inference and design updates with linear cost in the horizon; and
(iii) \emph{asymptotic consistency guarantees} for the resulting \gls*{eig} estimators when implemented with our approach.

\section{Background}
\label{sec:background}

\subsection{Sequential Bayesian Experimental Design}
\label{subsec:bed}

The goal of sequential \gls*{bed} is to select, at each experiment $t$, a design $\bxi_t$ that maximizes the expected information gain~\citep[EIG,][]{lindley1956measure} about the parameters:
$\bxi_t^\star = \arg\max_{\bxi_t \in \Omega} \; \mathcal{I}(\bxi_t)$, 
where $\Omega$ is the design space.  
The \gls*{eig} quantifies the reduction in posterior uncertainty about $\btheta$ after observing $\by_t$ under design $\bxi_t$, and is defined as:
\begin{align}\label{eq:eig}
\mathcal{I}&(\bxi_t) 
= \mathbb{E}_{p(\by_t \mid \bxi_t)}
   \!\bigg[\, \mathcal{H}[p(\btheta\!\mid\!h_{t-1})]
          - \mathcal{H}[p(\btheta\!\mid\!h_t)] \bigg] \\
&= \mathbb{E}_{p(\btheta \mid h_{t-1})\,p(\by_t \mid \btheta,\bxi_t)}
   \!\big[ \log p(\by_t\!\mid\!\btheta,\bxi_t) 
          - \log p(\by_t\!\mid\!\bxi_t) \big],\nonumber
\end{align}
with $h_{t-1} \!=\! \{\bxi_{1:t-1},\by_{1:t-1}\}$ the history, 
$p(\by_t \!\mid\! \btheta,\bxi_t)$ the likelihood,  
$p(\by_t \!\mid\! \bxi_t)$ the marginal predictive distribution,  
and $\mathcal{H}$ denoting entropy.  

In many applications, however, observations $\by_t$ are not generated directly by $\btheta$, but through an underlying dynamical system with latent states $\bx_{0:t}$. Such cases are naturally modeled as \glspl*{ssm}, for which the likelihood is not available in closed form, rendering standard \gls*{bed} methods inapplicable.

\subsection{State-Space Models (SSMs)} \label{subsec:ssm}

We focus on dynamical systems that can be described by Markovian \glspl*{ssm}. At each experiment or time step $t \!\in\! \naturals$, the system state $\bx_t \!\in\! \reals^{d_x}$ evolves and produces an observation $\by_t \!\in\! \reals^{d_y}$:
\begin{eqnarray}
        \bx_t &\sim& f(\bx_t\!\mid \! \bx_{t-1}, \btheta, \bxi_t), \label{eq:state}\\
        \by_t &\sim& g(\by_t\!\mid \! \bx_t, \btheta, \bxi_t), \label{eq:obs}
\end{eqnarray}
where $\btheta \!\in\! \reals^{d_\theta}$ are \emph{unknown} parameters and $\bxi_t \!\in\! \Omega \!\subset\! \reals^{d_\xi}$ is the design variable. 
Here, $f(\bx_t \!\mid\! \bx_{t-1}, \btheta, \bxi_t)$ is the transition density of the latent states, and $g(\by_t \!\mid\! \bx_t, \btheta, \bxi_t)$ is the observation model, both assumed to be differentiable \gls*{wrt} $\bxi_t$.
We assume that $\by_t$ (conditional on the state, parameters, and design) is conditionally independent of all other observations, and the prior \glspl*{pdf} of the state and the parameters, $p(\btheta)$ and $p(\bx_0)$, are known and independent.

\subsection{Particle Filters (PFs)} \label{subsec:pf}

In \glspl*{ssm}, the likelihood $p(\by_t\!\mid\!\btheta,\bxi_t,h_{t-1})$ typically has no closed form, as it requires marginalising the latent state,
\begin{equation}\label{eq:likelihood}
p(\by_t\! \mid \!\btheta,\!\bxi_t,\!h_{t-1})
\!=\!
\int \!g(\by_t \!\mid \!\bx_t,\btheta,\bxi_t)\,
p(\mathrm{d}\bx_t \!\mid \!\by_{1:t}, \btheta, \bxi_{1:t}).
\end{equation}
Particle filters \citep[PFs,][]{Gordon93,Doucet00,Doucet01,Djuric03} provide a standard solution by recursively approximating the filtering or posterior distribution with a weighted set of $M$ particles,
\[
p(\mathrm{d}\bx_t\!\mid \! \by_{1:t}, \btheta, \bxi_{1:t}) 
\approx \sum_{m=1}^M w_{\bx,t}^{(m)}\,\delta_{\bx_t^{(m)}}(\mathrm{d}\bx_t).
\]
At each time step, particles $\bx_t^{(m)}$ are propagated forward using the state transition $f$, while their weights $w_{\bx,t}^{(m)}$ are updated in proportion to the observation model $g$. This allows the algorithm to adaptively focus on regions of high likelihood and provides a flexible representation of complex, nonlinear distributions.

Standard PFs assume that both the parameters $\btheta$ and the designs $\bxi_{1:t}$ are known and fixed, so that only the latent states need to be inferred. In sequential \gls*{bed}, however, the parameters $\btheta$ are also unknown and must be estimated. This motivates the use of NPFs, which extend the PF framework to incorporate parameter inference.

\subsection{Nested Particle Filters (NPFs)} \label{subsec:npf}

NPFs \citep{Crisan18bernoulli} extend PFs to jointly estimate states and parameters \emph{online}, targeting the joint posterior
$p(\mathrm{d}\bx_{0:t}, \mathrm{d}\btheta\! \mid\! \by_{1:t}, \bxi_{1:t})$.
An NPF employs two intertwined layers of PFs:
(i) an \emph{inner} layer that estimates the state distribution conditional on each parameter particle
\[
p(\mathrm{d}\bx_t\!\mid \! \by_{1:t}, \btheta, \bxi_{1:t}) 
\approx \sum_{m=1}^M \sum_{n=1}^N w_{\btheta,t}^{(m)} w_{\bx,t}^{(m,n)}\,\delta_{\bx_t^{(m,n)}}(\mathrm{d}\bx_t),
\]
and
(ii) an \emph{outer} layer that represents the parameter posterior
\[
p(\mathrm{d}\btheta\!\mid \! \by_{1:t}, \bxi_{1:t}) 
\approx \sum_{m=1}^M  w_{\btheta,t}^{(m)}\,\delta_{\btheta_t^{(m)}}(\mathrm{d}\btheta).
\]
The parameter weights are updated using an estimate of the likelihood in \eqref{eq:likelihood} obtained from the inner filters.

This nested structure is related to sequential Monte Carlo squared \citep[SMC$^2$,][]{chopin2013smc2}, but differs in a key aspect. 
In SMC$^2$, when parameters are moved (e.g., via MCMC rejuvenation),
the inner filters are typically re-run using the full data history, leading to a cost that grows quadratically with time. This becomes problematic in long-horizon settings, both in runtime and memory.
NPFs instead use a \emph{jittering} kernel to perturb parameter particles,
\[
\btheta_t^{(m)} \sim \kappa_{M}\big(\mathrm{d}\btheta\! \mid\! \btheta_{t-1}^{(m)}\big),
\]
with a variance chosen to satisfy the regularity conditions required for the NPF convergence results
\citep{Crisan18bernoulli}, and tuned in practice to balance exploration and stability.
This controlled perturbation enables a purely online update: each inner PF is advanced a \emph{single} step (rather than reprocessing past data), yielding linear cost in $t$.

Several extensions replace the inner PF with more structured approximations (e.g., Gaussian or
Rao--Blackwellised variants) while retaining theoretical guarantees under modified assumptions
\citep{Perez-Vieites21,Fang23}.


%

\section{Proposed Method} \label{sec:method}

We propose a new sequential \gls*{bed} method for partially observable dynamical systems, deriving recursive estimators of the \gls*{eig} and its gradient. This enables online optimization in continuous design spaces, $\bxi_t \!\in\! \mathbb{R}^{d_\xi}$, using \gls*{sga}. We further establish consistency of the proposed \gls*{eig} estimator, and conclude the section by presenting the overall algorithmic scheme.

\subsection{EIG in SSMs}
\label{subsec:eig-ssm}

The \gls*{eig} expression in \eqref{eq:eig} shows the dependency on the likelihood $p(\by_t \!\mid\! \btheta,\bxi_t)$ and the marginal likelihood (evidence)
$p(\by_t \!\mid\! \bxi_t) = \mathbb{E}_{p(\btheta\mid h_{t-1})}[\,p(\by_t\!\mid\!\btheta,\bxi_t)]$.
Even when the likelihood is available in closed-form, estimating $\mathcal{I}(\bxi_t)$
is \emph{doubly intractable} because the evidence appears inside the outer
expectation that defines the utility, typically needing \gls*{nmc} methods or other approximations \citep{Rainforth18,Foster19}.

In partially observable dynamical systems, the difficulty compounds because both likelihood and evidence require marginalization over latent states.
For compactness we denote the likelihood and the evidence as
\begin{align}
L_{\btheta,\bxi_t}(\by_t)
&\coloneqq \mathbb{E}_{p(\bx_{0:t}\mid\btheta,h_{t-1})}\!\big[\, g(\by_t\!\mid\! \bx_t,\btheta,\bxi_t)\big],\nonumber
\\
Z_{\bxi_t}(\by_t)
&\coloneqq \mathbb{E}_{p(\btheta\mid h_{t-1})\,p(\bx_{0:t}\mid\btheta,h_{t-1})}\!\big[\, g(\by_t\!\mid\!\bx_t,\btheta,\bxi_t)\big],  \nonumber
\end{align}
respectively. Substituting these into \eqref{eq:eig} yields
\begin{equation}\label{eq:eig_partial}
  \mathcal{I}(\bxi_t) = \mathbb{E}_{p(\btheta \mid h_{t-1})\,\black{p(\by_t, \bx_{0:t} \mid \btheta, \bxi_t)}} 
  \left[\, \log \frac{L_{\btheta,\bxi_t}(\by_t)}{Z_{\bxi_t}(\by_t)} \right].
\end{equation}
This expression makes explicit that evaluating the \gls*{eig} requires not only the usual outer expectation over $\by_t$ (and $\btheta$), but also inner expectations over latent-state trajectories inside both $L_{\btheta,\bxi_t}$ and $Z_{\bxi_t}$. 
Although these quantities are analytically intractable, they admit natural Monte Carlo approximations: $L_{\btheta,\bxi_t}(\by_t)$ can be estimated by sampling state trajectories conditional on a fixed $\btheta$, while $Z_{\bxi_t}(\by_t)$ requires sampling jointly from the parameter prior and state dynamics. 
This compounded intractability motivates the gradient representation and efficient Monte Carlo estimators that we develop next.

\subsection{Gradient of the EIG}
\label{subsec:eig-grad}

Using Fisher's identity \citep{Douc14} on the \gls*{eig} of \eqref{eq:eig_partial}, we obtain a gradient representation that separates (i) derivatives of the likelihood and evidence and (ii) the design score at time $t$. Let
\begin{align}
\Gamma_{\bxi_t}(\by_t,\bx_{0:t},\btheta)
&\coloneqq g(\by_t\!\mid\! \bx_t,\btheta,\bxi_t)
f(\bx_t\!\mid\! \bx_{t-1},\btheta,\bxi_t)\,\times \nonumber\\
&\quad \times p(\bx_{0:t-1} \!\mid\! \btheta, h_{t-1}) p(\btheta \!\mid\! h_{t-1}) \nonumber
\end{align}
be the joint distribution of states, parameters, and observations given a design (generative model at time $t$). We assume the availability of an approximation to the joint posterior distribution of states and parameters at the previous time step, $p(\bx_{0:t-1}, \btheta \!\mid\! h_{t-1})$. 

The gradient of the \gls*{eig} can be written as
\begin{align} \label{eq:gradient_eig}
\nabla_{\bxi_t}\mathcal{I}(\bxi_t) &= \mathbb{E}_{\Gamma_{\bxi_t}}\!\Bigg[ \frac{\nabla_{\bxi_t} L_{\btheta,\bxi_t}(\by_t)}{L_{\btheta,\bxi_t}(\by_t)} - \frac{\nabla_{\bxi_t} Z_{\bxi_t}(\by_t)}{Z_{\bxi_t}(\by_t)} \nonumber \\
 &~ + \log\!\left[\,\frac{L_{\btheta,\bxi_t}(\by_t)}{Z_{\bxi_t}(\by_t)}\right] s_{\btheta,\bxi_t}(\bx_{t-1:t},\by_t) \Bigg],
\end{align}
where ${\Gamma_{\bxi_t}}$ stands for ${\Gamma_{\bxi_t}(\by_t,\bx_{0:t},\btheta)}$, and the design score at time $t$ is
\begin{align}
s_{\btheta,\bxi_t}(\bx_{t-1:t},\by_t)
&\coloneqq \nabla_{\bxi_t}\log g(\by_t\!\mid\! \bx_t,\btheta,\bxi_t)
\nonumber\\
&\quad + \nabla_{\bxi_t}\log f(\bx_t\!\mid\! \bx_{t-1},\btheta,\bxi_t), \nonumber
\end{align}
with any of the terms vanishing when they do not depend on $\bxi_t$. See details in Appendix~\ref{ap:gradient}.

Direct evaluation of \eqref{eq:gradient_eig} is intractable. Both the likelihood $L_{\btheta,\bx_t}(\by_t)$ and the evidence $Z_{\bx_t}(\by_t)$, as well as their gradients, are defined by expectations over latent-state trajectories. 
For instance, to sample a new state $\bx_t$ given a new $\btheta$, one needs the full path
\begin{align}
p(\bx_{0:t}\!&\mid\! \btheta,h_{t-1}) \nonumber \\
&\propto p(\bx_0)\prod_{s=1}^{t}
f(\bx_s\!\mid\!\bx_{s-1},\btheta,\bxi_s)\,\prod_{s=1}^{t-1}g(\by_s\!\mid\! \bx_s,\btheta,\bxi_s), \nonumber
\end{align}
which cannot be updated incrementally without revisiting all $s=1,\dots,t-1$. Naïvely, this yields $\mathcal{O}(t^2)$ cost across $t$ steps, as each new gradient evaluation requires resimulating and reweighting entire trajectories.

In the next subsection, we show how this quadratic bottleneck can be avoided using NPFs, yielding Monte Carlo gradient estimators that require only a \emph{single} forward propagation step per iteration, i.e., with complexity $\mathcal{O}(t)$.

\subsection{Monte Carlo Estimators using NPF}
\label{subsec:estimators}

We now instantiate the terms in \eqref{eq:gradient_eig} using the approximation of the posterior distributions available at time $t{-}1$, obtained with the NPF framework. Let
$\{\btheta_{t-1}^{(m)},\,w_{\btheta,t-1}^{(m)}\}_{m=1}^{M}$ denote the outer (parameter) particles and, for each $m$, let
$\{\bx_{0:t-1}^{(m,n)},\,w_{\bx,t-1}^{(m,n)}\}_{n=1}^{N}$ be the inner (state-trajectory) particles. This yields particle
approximations to $p(\btheta\!\mid\! h_{t-1})$ and $p(\bx_{0:t-1}\!\mid\! \btheta,h_{t-1})$ that we \emph{update online} without
reprocessing past data.

For approximating the \gls*{eig} or any other quantity in \eqref{eq:eig_partial} or \eqref{eq:gradient_eig} given a design $\bxi_t$, we first sample from the joint $\Gamma_{\bxi_t}(\by_t, \bx_{0:t}, \btheta)$. 
In the NPF framework, we 
reuse samples from the previous posterior and only draw
one-step predictive states and pseudo-observations
\begin{align}
\widetilde{\bx}_t^{(m,n)} &\sim f(\bx_t\mid \bx_{t-1}^{(m,n)},\btheta_{t-1}^{(m)},\bxi_t),\nonumber\\
\widetilde{\by}_t^{(m,n)} &\sim g(\by_t\mid \widetilde{\bx}_t^{(m,n)},\btheta_{t-1}^{(m)},\bxi_t).\nonumber
\end{align}
Thus, Monte Carlo samples from $\Gamma_{\bxi_t}$ form the array
$\{(\btheta_{t-1}^{(m)},\bx_{0:t-1}^{(m,n)},\widetilde{\bx}_t^{(m,n)},\widetilde{\by}_t^{(m,n)})\}$, needed for the \emph{outer} expectation in \eqref{eq:gradient_eig}, with weights
\[
w_{\by,t}^{(m,n)} = w_{\btheta,t-1}^{(m)} \, w_{\bx,t-1}^{(m,n)} .
\]
For notational clarity, we relabel the pseudo-observations as
$\widetilde{\by}_t^{(\ell)}$, $\ell = 1,\dots,L$, $L = M \times N$,
with corresponding weights $w_{\by,t}^{(\ell)}$, where each index $\ell$ corresponds to a unique pair $(m,n)$.

The samples required for the \emph{inner} expectations that integrate out latent states are sampled in a similar way.
We then sample the next state in the trajectory with different conditioning:
(i) for $L_{\btheta,\bxi_t}(\widetilde{\by}_t)$ we \emph{fix} the parameters, $\btheta_{t-1}^{(m)}$; and (ii) for $Z_{\bxi_t}(\by_t)$ we sample new parameters in order to draw new states. With the NPF particles, we can operate:

\emph{(a) For }$L_{\btheta,\bxi_t}(\by_t)$: fix $\btheta=\btheta_{t-1}^m$ and propagate the associated inner particles one step,
\[
\ddot{\bx}_t^{(m,j)} \sim f(\bx_t\!\mid \!\bx_{t-1}^{(m,j)},\btheta_{t-1}^{(m)},\bxi_t),
\]
to obtain the Monte Carlo approximation
\begin{equation}
\label{eq:likelihood_estimator}
\widehat{L}_{\btheta_{t-1}^{(m)},\bxi_t}^N(\by_t)
= \sum_{j=1}^N w_{\bx,t-1}^{(m,j)}\, g(\by_t \!\mid\! \ddot{\bx}_t^{(m,j)},\btheta_{t-1}^{(m)},\bxi_t).
\end{equation}

\emph{(b) For }$Z_{\bxi_t}(\by_t)$: jitter parameter particles and propagate one step,
\begin{align}
\dot{\btheta}_{t-1}^{(i)} &\sim \kappa_M(\cdot\!\mid\! \btheta_{t-1}^{(i)}),
\nonumber\\
\dot{\bx}_t^{(i,j)} &\sim f(\bx_t\!\mid\! \bx_{t-1}^{(i,j)},\dot{\btheta}_{t-1}^{(i)},\bxi_t), \nonumber
\end{align}
then average across both indices with the outer and inner weights,
\begin{equation}
\label{eq:evidence_estimator}
\widehat{Z}_{\bxi_t}^{M,N}(\by_t)
= \sum_{i=1}^M \sum_{j=1}^N w_{\btheta,t-1}^{(i)} w_{\bx,t-1}^{(i,j)}\,
    g(\by_t \!\mid\! \dot{\bx}_t^{(i,j)},\dot{\btheta}_{t-1}^{(i)},\bxi_t).
\end{equation}

Thus, to approximate the \gls*{eig} in \eqref{eq:eig_partial} we draw:
(i) parameters $\{\btheta_{t}^{(m)}, \dot{\btheta}_{t}^{(m)}\}$ for $m=1,\dots,M$; 
(ii) conditional state trajectories $\{ \bx_{0:t-1}^{(m,n)}, \widetilde{\bx}_t^{(m,n)}, \dot{\bx}_t^{(m,n)}, \ddot{\bx}_t^{(m,n)} \}$ for $n=1,\dots,N$; 
and (iii) pseudo-observations $\widetilde{\by}_t^{(\ell)} \!\sim\! g(\by_t\! \mid\! \widetilde{\bx}_t^{(m,n)},\btheta^{(m)},\bxi_t)$, for $\ell=1,\dots,L$.
This yields the \gls*{nmc} approximation
\begin{equation} \label{eq:eig_estimator}
\widehat{\mathcal{I}}(\bxi_t) 
=   \sum_{\ell=1}^L
    \black{w_{\by,t-1}^{(\ell)}}
   \log \frac{\widehat{L}_{\btheta,\bxi_t}^N(\widetilde{\by}_t^{(\ell)})}
             {\widehat{Z}_{\bxi_t}^{M,N}(\widetilde{\by}_t^{(\ell)})}.
\end{equation}

Following a similar procedure, the gradients $\nabla_{\bxi_t} L_{\btheta,\bxi_t}(\by_t)$ and $\nabla_{\bxi_t} Z_{\bxi_t}(\by_t)$ of \eqref{eq:gradient_eig} can be approximated. See details in Appendix~\ref{ap:approximations}.

\subsection{Consistency of the EIG Estimator}
\label{subsec:consistency}

Consistency of our estimator \(\widehat{\mathcal{I}}(\bxi_t)\) follows from two ingredients: (i) NPF convergence under standard conditions (small/regular jittering kernels, Lipschitz state posterior in $\btheta$, and a bounded positive likelihood), and (ii) \gls*{nmc} consistency provided the \gls*{eig} integrand, $\log(L/Z)$, is Lipschitz and square-integrable. Formal statements of these assumptions are given in Appendix~\ref{ap:consistency}.

\begin{theorem}\label{th:bigtheorem}
Let $\widehat{\mathcal{I}}(\bxi_t)$ denote the \gls*{nmc} estimator of the \gls*{eig} \eqref{eq:eig_estimator}, constructed using $M$ parameter particles, $N$ state particles per parameter, and $L$ pseudo-observations.
Under Assumptions~\ref{as:kernel scaling}–\ref{as:nmc} (in Appendix~\ref{ap:consistency}), for any $t$ and $\bxi_t$
\[
\widehat{\mathcal{I}}(\bxi_t) \;\xrightarrow[L,M,N \to \infty]{\text{a.s.}}\;  \mathcal{I}(\bxi_t).
\]
\end{theorem}

The proof combines \gls*{nmc} results~\citep{Rainforth18}, ensuring consistency of nested expectations under Assumption~\ref{as:nmc}, with NPF convergence guarantees \citep{Crisan18bernoulli}, which establish almost-sure convergence of empirical state–parameter measures under Assumptions~\ref{as:kernel scaling}–\ref{as:likelihood}. Together, these results imply that the Monte Carlo averages in our estimator converge almost surely to the true \gls*{eig}. Full details are given in Appendix~\ref{ap:consistency}.

\subsection{Overall Algorithm}
\label{subsec:algorithm}

BAD-PODS ({B}ayesian {A}daptive {D}esign for {P}artially {O}bservable {D}ynamical {S}ystems), selects designs sequentially at each time step $t$. 
Given the current history $h_{t-1}$, we optimise $\bxi_t$ via \gls*{sga} (Section~\ref{subsec:eig-grad}), using the gradient estimator of the \gls*{eig} computed with \gls*{nmc} samples from $\Gamma_{\bxi_t}$ (Section~\ref{subsec:estimators}). After $K$ optimisation steps, with step sizes $\eta_k$, we obtain the optimized design $\bxi_t^\star$ and the real observation $\by_t$ is collected. The posterior is then updated via an NPF (Section~\ref{subsec:npf}): parameter particles are jittered, states are propagated through $f$, and weights are adjusted using $g$, with resampling to maintain diversity. This recursive construction avoids reprocessing past data, with a per-step cost of $\mathcal{O}((KL+1)MN)$ (and $\mathcal{O}(KLMNt)$ over $t$ steps), ensuring linear scaling in time while supporting gradient-based design optimization.

The algorithm takes as input the prior distributions, particle numbers $(M,N)$, and optimization hyperparameters $(K,\eta_k)$, and produces as output the optimized design sequence $\{{\bxi}_t^\star\}_{t=1:T}$ together with posterior approximations of states and parameters. 
See Appendix~\ref{ap:algorithm} for full implementation details.

\section{Related Work}\label{sec:related work}

A central challenge in \gls*{bed} is the cost and complexity of estimating the \gls*{eig} (or its gradient) \citep{Rainforth24,huan2024optimal}. For models with tractable likelihoods $p(\by\!\mid\!\btheta,\bxi)$, the \gls*{nmc} estimator has been studied in depth \citep{Rainforth18}, with subsequent work proposing variational formulations and bounds to improve convergence \citep{Foster19,Foster20,Foster21}. These approaches also enable differentiable objectives for gradient-based design in continuous spaces. However, they rely on explicit pointwise likelihood evaluation and do not extend directly to partially observable dynamical systems, where likelihoods require marginalization over latent states.

When likelihoods are implicit (intractable pointwise but possible to simulate from), likelihood-free design methods replace direct evaluation by surrogates or simulation-based objectives. Examples include variational mutual-information bounds and amortized estimators \citep{Kleinegesse20,Kleinegesse21,Ivanova21}, density-ratio or classifier-based estimates of information gain \citep{Kleinegesse19}, and \gls*{abc}-style utilities \citep{Dehideniya18,Drovandi13,Hainy16,Price16}. While effective in simulator settings, most such methods target \emph{static} design (optimize once, then deploy) rather than fully sequential/adaptive \gls*{bed}. Moreover, surrogate-based adaptive variants \citep[e.g.,][]{Ivanova21} typically require an offline dataset to train an implicit likelihood, which may be unavailable or quickly misspecified in evolving dynamical systems.

Design for dynamical systems has also been studied under full state observability. 
\citet{Iqbal24,iqbal2024recursive} consider adaptive design when the state is fully observed, making the likelihood tractable. 
Both papers leverage particle methods for joint state–parameter inference: \citet{Iqbal24} use a reversed (“inside-out”) SMC$^2$, and \citet{iqbal2024recursive} adopt NPF's jittering to rejuvenate parameter particles. 
However, these recent advances in \gls*{bed} for dynamical systems assume full state observability and do not address the latent-state marginalizations required in partially observable \glspl*{ssm}.

\section{Experiments}\label{sec:experiments}

We evaluate our method on two partially observable dynamical systems. Since no existing methods target this setting, we compare against {three} standard \gls*{bed} baselines~\citep{foster2021thesis}:
(i) \emph{random} designs sampled uniformly from $\Omega$, 
(ii) an \emph{oracle} version of BAD-PODS, which (instead of gradient-based optimisation) evaluates the estimated \gls*{eig} on a dense discretisation of $\Omega$ and selects the best grid point at each step (providing an approximate upper bound); and 
(iii) \emph{static} \gls*{bed}, a non-adaptive version of our method where the full sequence $\bxi_{1:T}$ is optimized offline. Unlike our sequential approach, the static baseline cannot adapt as data arrive.

Performance is measured by the \emph{total \gls*{eig}} (TEIG), the accumulated information gain across time. We also report \emph{relative improvements}, 
\[
\Delta\text{TEIG}^{\text{(baseline)}} = \sum_{s=1}^t \bigg(\widehat{\mathcal{I}}(\bxi_s^{\text{BAD-PODS}}) - \widehat{\mathcal{I}}(\bxi_s^{\text{baseline}}) \bigg),
\] 
against each baseline to highlight adaptive gains. Results are averaged over 50 seeds with bootstrap 95\% confidence intervals.
%

{Python code to reproduce all experiments is released at \url{https://anonymous.4open.science/r/badpods-2C8D}. Details of the computing infrastructure are provided in Appendix~\ref{ap:infrastructure}.}

\subsection{Two-group SIR Model}\label{subsec:sir}

The \gls*{sir} model is a standard testbed in \gls*{bed}, but prior work often assumes fully observed states \cite{Ivanova21} or static design \cite{Kleinegesse19}. Here we use a \emph{partially observable}, stochastic two-group \gls*{sir} model with \emph{online} sequential design. The latent state tracks the susceptible/infectious counts in each group \(\mathbf{X}^{(g)}(\tau) = \big(S^{(g)}(\tau), I^{(g)}(\tau)\big)^\top\), and the unknown parameters are the infection and recovery rates $\btheta=\{(\beta^{(g)},\gamma^{(g)})\}_{g=1}^2$. At each time step \(t\), the design \(\bxi_t=(\xi_t^{(1)},\xi_t^{(2)})\) allocates a fixed sampling effort across the two groups (simplex constraint), producing noisy incidence observations. The design therefore modulates the observation process, while the epidemic dynamics remain unchanged. Full model details are given in Appendix~\ref{ap:sir}.

\paragraph{Experimental setup.}
We set the horizon to $T\!=\!200$ and simulate ground-truth trajectories from the two-group \gls*{sir} model with group sizes $N^{(1)}\!=\!N^{(2)}\!=\!200$ and initial infections $I_{0}^{(1)}\!=\!I_{0}^{(2)}\!=\!5$. The cross-group mixing matrix $M$, detection scales $\rho^{(g)}$, sampling budget $\kappa$, and the parameters of the second group $(\beta^{(2)},\gamma^{(2)})$ are fixed and known. Inference targets only the first-group parameters $(\beta^{(1)},\gamma^{(1)})$. At each time step, the design $\bxi_t$ allocates observation effort between $y_t^{(1)}$ and $y_t^{(2)}$.
%
For more details see Appendix~\ref{ap:sir}.

\paragraph{Experimental results.}
Table~\ref{tab:teig} reports TEIG for all four methods. BAD-PODS consistently performs close to the oracle solution, and outperforms both static and random baselines, with gaps widening over time. Results for the static method are truncated as optimization becomes infeasible at longer horizons.

\begin{table}[t]
\caption{TEIG across time for the two-group \gls*{sir} model (top), moving-source model (middle) and ecological growth model (bottom). Means over 50 seeds with 95\% \gls*{bca} CIs. Static results truncated at longer horizons due to computational cost.}
\label{tab:teig}
\centering
\scriptsize
\setlength{\tabcolsep}{5pt}
\renewcommand{\arraystretch}{0.9}
\resizebox{\columnwidth}{!}{%
\begin{tabular}{rcccc}
\toprule
\multicolumn{5}{c}{\textbf{Two-group \gls*{sir} model}} \\[0.4ex]
$t$ & Oracle & BAD-PODS & Random & Static \\
\midrule
50  & 1.450 & \textbf{1.390} & 1.364 & 1.167 \\
    & [1.376,1.520] & [1.324,1.460] & [1.287,1.445] & [1.117,1.220] \\
100 & 3.330 & \textbf{3.283} & 3.152 & 2.529 \\
    & [3.215,3.442] & [3.173,3.380] & [3.040,3.268] & [2.444,2.621] \\
150 & 4.848 & \textbf{4.748} & 4.518 & 3.547 \\
    & [4.747,4.959] & [4.609,4.871] & [4.385,4.627] & [3.454,3.658] \\
200 & 6.043 & \textbf{5.918} & 5.621 & -- \\
    & [5.939,6.168] & [5.765,6.055] & [5.466,5.793] & \\
\midrule
\multicolumn{5}{c}{\textbf{Moving source model}} \\[0.4ex]
$t$ & Oracle & BAD-PODS & Random & Static \\
\midrule
10 & 0.304 & \textbf{0.297} & 0.242 & 0.270 \\
   & [0.296,0.312] & [0.289,0.308] & [0.222,0.263] & [0.258,0.284] \\
20 & 0.698 & \textbf{0.688} & 0.582 & 0.570 \\
   & [0.672,0.731] & [0.661,0.726] & [0.531,0.649] & [0.541,0.604] \\
30 & 1.158 & \textbf{1.130} & 0.940 & -- \\
   & [1.103,1.223] & [1.075,1.197] & [0.855,1.044] & \\
40 & 1.515 & \textbf{1.465} & 1.217 & -- \\
   & [1.451,1.591] & [1.400,1.556] & [1.136,1.344] & \\
50 & 1.805 & \textbf{1.733} & 1.435 & -- \\
   & [1.732,1.888] & [1.658,1.840] & [1.354,1.555] & \\
   \midrule
\multicolumn{5}{c}{\textbf{Ecological growth model}} \\[0.4ex]
$t$ & Oracle & BAD-PODS & Random & Static \\
\midrule
5 & 0.935 & \textbf{0.902} & 0.725 & 0.897	 \\
   & [0.896, 0.970] & [0.854, 0.946] & [0.657, 0.790] & [0.866, 0.946]	 \\
10 & 1.618 & \textbf{1.579} & 1.337 & 1.571 \\
   & [1.554, 1.690] & [1.499, 1.650] & [1.245, 1.429] & [1.507, 1.627] \\
15 & 2.277 & \textbf{2.215} & 1.972 & -- \\
   & [2.175, 2.379] & [2.113, 2.312] & [1.844, 2.109] & \\
20 & 2.621 & \textbf{2.545} & 2.344 & -- \\
   & [2.492, 2.741] & [2.422, 2.673] & [2.211, 2.499] & \\
\bottomrule
\end{tabular}%
}
\end{table}

\begin{figure}[t]
\includegraphics[width=0.47\textwidth]{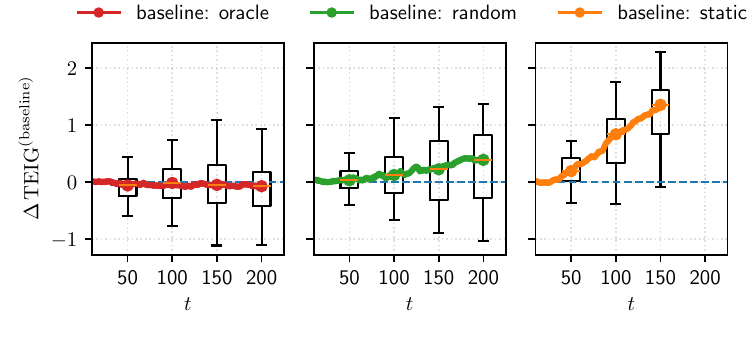}
\caption{$\Delta$TEIG for the two-group \gls*{sir}. Boxplots (median and interquartiles) compare BAD-PODS with the oracle (left, red), random designs (middle, green) and static \gls*{bed} (right, orange). Higher values indicate better performance for BAD-PODS.}
\label{fig:sir-diffs}
\end{figure}

\begin{figure}[t]
\centering
\includegraphics[width=0.32\textwidth]{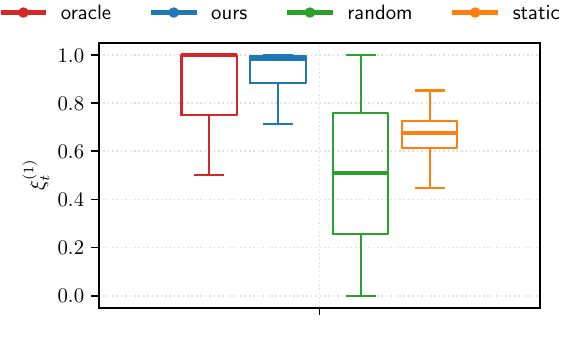}
\caption{Distribution of design component $\xi_t^{(1)}$ across seeds and time (boxplots: median and interquartiles). BAD-PODS allocates more effort to group~1, which contains the unknown parameters.}
\label{fig:sir-designs}
\end{figure}

Figure~\ref{fig:sir-diffs} plots $\Delta$TEIG, confirming that gains accumulate over time. The improvements of BAD-PODS relative to the static and random baselines grow steadily with time, and are particularly higher with respect to the static baseline. 
Figure~\ref{fig:sir-designs} further explains this behavior: while random designs are highly variable, both static and sequential approaches converge to structured allocations, with BAD-PODS consistently allocating more effort to group~1 (the group with unknown parameters).

\paragraph{Runtime and memory.}
For the \gls*{sir} experiment with $T\!=\!150$, average wall-clock runtimes are $2.09$ minutes for the random baseline (no optimisation), $31.05$ for the static baseline (single high-dimensional optimisation over $\bxi_{1:T}$), and $12.39$ for BAD-PODS (per-step optimisation with online inference), yielding a $\approx\!2.5\times$ speed-up over the static approach. Memory is an equally important bottleneck: at $T\!=\!150$, the static baseline uses $\approx 25$\,GB GPU and $\approx 25$\,GB CPU memory, whereas BAD-PODS uses $\approx 10$\,GB GPU and $\approx 1$\,GB CPU. This gap grows with $T$ since BAD-PODS only needs to store the current $M\!\times\! N$ particle set, while the static baseline must store and optimise over the full design tensor and latent trajectories of size $M\!\times\! N\!\times\! T$, making long horizons increasingly expensive.

\subsection{Moving Source Location}
\label{subsec:source}



Source localisation is a standard \gls*{bed} testbed \citep{Ivanova21,Foster21}, typically studied with a static source and sensor positions as designs. Here, we use a more challenging \emph{moving} source, with designs given by the \emph{orientations} of fixed sensors. This setup is closely related to sensor-control/target-tracking, where sensor configurations are adapted online to improve \emph{state estimation} \citep{koch2016tracking}. 

A latent source state \(\bx_t\!=\!(p_{x,t},p_{y,t},\phi_t)^\top\) evolves according to a motion model with unknown parameters \(\btheta\!=\!(v_x,v_y,v_\phi)\). At each time \(t\), the design \(\bxi_t\!\in\![-\pi,\pi)^J\) selects the orientation of $J$ sensors, and each sensor returns a noisy intensity whose mean depends on distance and a directional gain term that peaks when the sensor points toward the source. Hence, \(\bxi_t\) affects only the observation model. Full dynamics and observation model details are in Appendix~\ref{ap:source}.

\begin{figure}[t]
\centering
\includegraphics[width=0.3\textwidth]{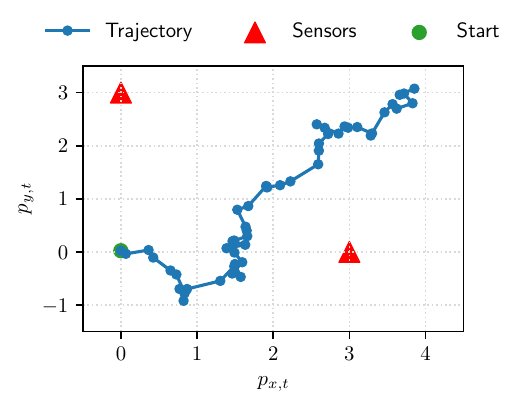}
\caption{Example trajectory of the moving source for $T\!=\!50$ (blue line) together with fixed sensor locations (red triangles). 
}
\label{fig:source-trajectory}
\end{figure}

\paragraph{Experimental setup.}
We simulate a moving source over a horizon of $T\!=\!50$ steps starting at position $\bx_0\! =\! (0,0,0)$. The source parameters to be inferred are the horizontal and vertical velocities, $\btheta=(v_x,v_y)$, while the angular velocity $v_\phi$ and all other observation parameters are treated as known.
We deploy $J\!=\!2$ fixed sensors located at $\bs_1\!=\!(0,3)$ and $\bs_2\!=\!(3,0)$, each with orientation design variables $\bxi_t\!=\!(\xi_{t,1},\xi_{t,2}) \in [-\pi,\pi)^2$. Figure~\ref{fig:source-trajectory} illustrates the setup and one example trajectory. 
%
Additional implementation details and parameter values are given in Appendix~\ref{ap:source}.

\begin{figure}[t]
\includegraphics[width=0.47\textwidth]{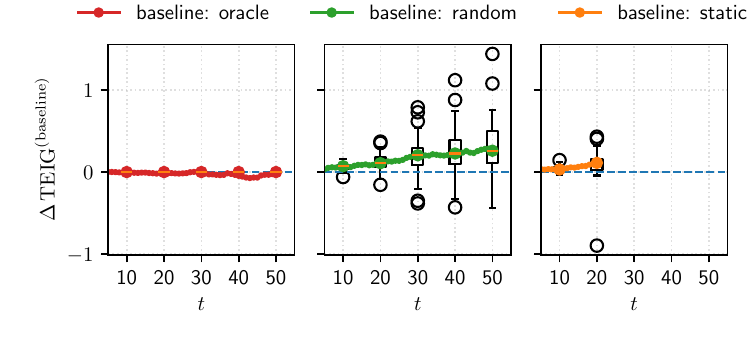}
\caption{
$\Delta$TEIG for the moving source location. Boxplots (median and interquartiles) compare BAD-PODS with oracle (left, dark red), random (middle, green) and static baselines (right, orange). Higher values indicate better performance for BAD-PODS.
}
\label{fig:source-diffs}
\end{figure}

\paragraph{Experimental results.}
Table~\ref{tab:teig} reports the TEIG for our sequential method compared against the oracle, and the random and static baselines. Once again, our approach achieves the highest TEIG across all horizons, performing very close to the oracle and confirming the benefits of adapting designs online in partially observable settings.
Figure~\ref{fig:source-diffs} illustrates these results more clearly by showing boxplots of TEIG differences relative to each baseline. The performance of BAD-PODS is very close to the oracle (almost flat), and the improvements \gls*{wrt} the other baselines accumulate steadily over time as information gain compounds. As in the \gls*{sir} experiment, results for the static method are truncated at larger horizons due to computational infeasibility of offline optimization in high-dimensional design spaces.

\begin{figure}[t]
\centering
\includegraphics[width=0.38\textwidth]{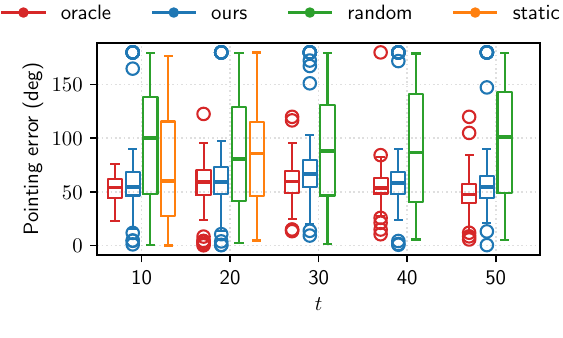}
\caption{Pointing error (in degrees) of sensor orientations relative to the target at selected time steps. Boxplots (median and interquartiles) show the distribution across seeds and sensors for the oracle, our method, random design, and static \gls*{bed} approach. Lower values correspond to higher orientation accuracy.}
\label{fig:pointing-error}
\end{figure}

Finally, we assess the quality of the selected sensor orientations in terms of their pointing error, i.e., the angular deviation between a sensor’s chosen orientation and the true bearing to the source.
Figure~\ref{fig:pointing-error} shows boxplots of pointing error distributions at selected time steps across time, seeds and sensors.
Both random and static baselines exhibit wide variability, often pointing far away from the target.
In contrast, our sequential method consistently achieves much smaller pointing errors, similar to those obtained with the oracle, demonstrating its ability to align sensors toward the target more frequently and with greater accuracy.
However, perfect accuracy is not achieved since outliers are expected in this experiment. This is due to the signal-to-noise ratio, which sometimes is intrinsically low (e.g., the source is far and observations are dominated by background/noise), so pointing becomes weakly identifiable. 

\subsection{Ecological Growth Model}
\label{subsec:growth}

We next consider a growth/harvest \gls*{ssm}, commonly used in ecology to represent the evolution of a population under harvesting pressure \citep{zhou2009modified}. Unlike the previous examples, where the design enters only through the observation model, here, the design affects \emph{both} the transition dynamics and the observation process, yielding a more general testbed for sequential \gls*{bed}.

The latent state \(x_t\!\in\!\mathbb{R}_+\) denotes the (scaled) population size at time \(t\), and the design \(\xi_t\! \in \![0,1]\) controls the harvesting intensity/effort (e.g., fishing effort or trapping rate). Observations provide a noisy, partially saturated measurement of the harvested amount. Full model details are in Appendix~\ref{ap:growth}.
%



\paragraph{Experimental setup.}
We simulate the growth/harvest model over a horizon of \(T=20\). The initial latent state is set to $x_0 = 0.4k$, where $r\!=\!0.5$ and $k\!=\!300$. 
The parameters to be inferred are the intrinsic growth rate and the carrying capacity $\btheta\!=\!(r,k)$, and all other observation parameters are treated as known.
Additional implementation details and parameter values are provided in Appendix~\ref{ap:growth}.

\paragraph{Experimental results.}
Table~\ref{tab:teig} reports TEIG for all four methods. As in the other models, BAD-PODS consistently performs close to the oracle, and outperforms both static and random baselines. Results for the static method are truncated as optimization becomes infeasible at longer horizons.

\begin{figure}[t]
\includegraphics[width=0.47\textwidth]{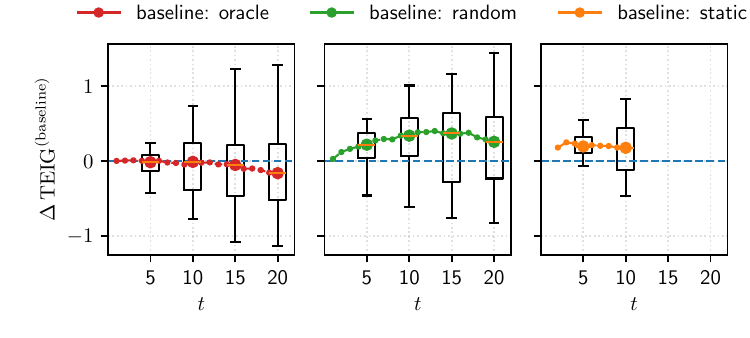}
\caption{$\Delta$TEIG for the ecological growth model. Boxplots (median and interquartiles) compare BAD-PODS with oracle (left, red), random (middle, green) and static baselines (right, orange). Higher values indicate better performance for BAD-PODS.}
\label{fig:growth-diffs}
\end{figure}

Figure~\ref{fig:growth-diffs} shows relative improvements in accumulated information, $\Delta\mathrm{TEIG}$, against each baseline. BAD-PODS achieves positive gains over random designs across the horizon, with the improvement tending to plateau toward $T=20$. This is expected: once the posterior concentrates, the marginal information gain per additional step diminishes, and the best designs become less distinguishable. BAD-PODS shows a smaller gap or improvement \gls*{wrt} the static baseline, since a one-shot static optimisation can already recover a reasonable policy when $T$ is short.

\begin{figure}[t]
\centering
\includegraphics[width=0.32\textwidth]{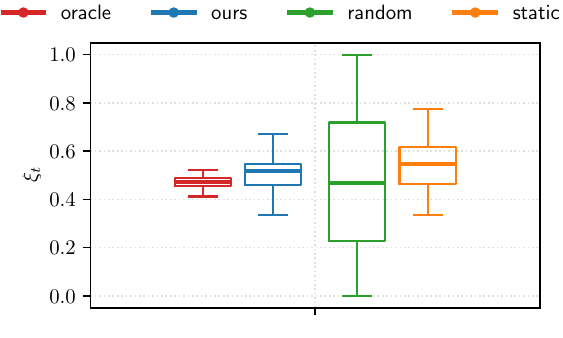}
\caption{Distribution of design $\xi_t$ across seeds and time (boxplots: median and interquartiles) for the ecological growth model.}
\label{fig:growth-designs}
\end{figure}

Figure~\ref{fig:growth-designs} reports the distribution of selected designs $\xi_t$ (aggregated over seeds and time). Both the oracle (grid search) and BAD-PODS concentrate around an intermediate harvesting effort ($\xi_t \!\approx\! 0.5$), while random designs spread across the full interval. This ``sweet spot'' arises from a trade-off: small $\xi_t$ produces weak harvest signals (low signal-to-noise, hence low information), whereas large $\xi_t$ quickly depresses the population state $x_t$ and drives the observation into a saturated/low-sensitivity regime.

\section{Conclusions}\label{sec:conclusion}
\glsresetall

We introduced an online sequential \gls*{bed} method for partially observable dynamical systems, modeled as \glspl*{ssm}. Our approach combines two key ingredients: (i) new estimators of the expected information gain and its gradient that explicitly account for latent states, and (ii) an NPF-based construction that reuses state–parameter particles to evaluate these quantities online. This yields an algorithm whose cost grows only linearly with time, avoids reprocessing past data, and remains asymptotically consistent under mild assumptions. The method supports continuous design spaces and stochastic-gradient optimization, making it broadly applicable. Through experiments on a multi-group epidemiological model, a dynamical source–tracking problem and an ecological growth model, we demonstrated that online adaptation significantly improves information efficiency over static \gls*{bed} or random design baselines.

\section*{Impact Statement}


This paper contributes algorithms and theory for Bayesian experimental design. The methods are broadly applicable and may reduce the number of experiments needed in scientific and engineering workflows. Any societal consequences depend on the application, and we do not claim specific impacts beyond the methodological scope of this work.

\bibliography{bibliography}
\bibliographystyle{icml2026}

\newpage
\appendix
\onecolumn
\section{Derivation of the EIG Gradient} 
\label{ap:gradient}

We derive a gradient of the \gls*{eig} \gls*{wrt} the design $\bxi_t$. The design affects both the transition and observation models, 
\(f(\bx_t\!\mid\!\bx_{t-1},\btheta,\bxi_t)\) and \(g(\by_t\!\mid\!\bx_t,\btheta,\bxi_t)\).
For clarity, we introduce the joint distribution,
\[
\Gamma(\by_t,\bx_{0:t},\btheta\! \mid\! \bxi_t,h_{t-1})
= p(\btheta\!\mid\!h_{t-1})\,
  p(\bx_{0:t-1}\!\mid\!\btheta,h_{t-1})\,
  f(\bx_t\!\mid\!\bx_{t-1},\btheta,\bxi_t)\,
  g(\by_t\!\mid\!\bx_t,\btheta,\bxi_t).
\]
Using this notation, the \gls*{eig} in \eqref{eq:eig_partial} (Section~\ref{subsec:eig-ssm}) can be compactly expressed as
\[
\mathcal{I}(\bxi_t)
= \mathbb{E}_{\Gamma(\cdot \mid \bxi_t,h_{t-1})}
\!\left[\log\frac{L_{\btheta,\bxi_t}(\by_t)}{Z_{\bxi_t}(\by_t)}\right],
\qquad
L_{\btheta,\bxi_t}(\by_t)=p(\by_t\!\mid\!\btheta,\bxi_t),~~
Z_{\bxi_t}(\by_t)=p(\by_t\!\mid\!\bxi_t).
\]

To compute its gradient, we write the expectation explicitly in integral form:
\begin{align}
\nabla_{\bxi_t}\mathcal{I}(\bxi_t)
&= \nabla_{\bxi_t}\!\!
\int\!\!\int\!\!\int
p(\btheta\!\mid\! h_{t-1})\,
p(\bx_{0:t-1}\!\mid\!\btheta, h_{t-1})\,
f(\bx_t\!\mid\!\bx_{t-1}, \btheta, \bxi_t)\,
g(\by_t\!\mid\!\bx_t, \btheta, \bxi_t) \,\times
\nonumber\\[-2pt]
&\hspace{2.3cm}\times\,
\bigg(\,
\log p(\by_t\!\mid\!\btheta, \bxi_t)
\,-\, \log p(\by_t\!\mid\!\bxi_t)
\bigg)
\,
\mathrm{d}\by_t\,\mathrm{d}\bx_{0:t}\,\mathrm{d}\btheta.
\nonumber
\end{align}

Applying the product rule yields three contributions: from \(f\), from \(g\), and from the log–ratio term:
\begin{align}
    \nabla_{\bxi_t}\!\mathcal{I}(\bxi_t) 
    &= 
    \int \int \int p(\btheta\!\mid\! h_{t-1})   p(\bx_{0:t-1}\!\mid \!\btheta, h_{t-1})  \Big(\, \nabla_{\bxi_t} f(\bx_t \!\mid\! \bx_{t-1}, \btheta, \bxi_t)\Big)\, g(\by_t \!\mid\! \bx_t, \btheta, \bxi_t) \,\times \nonumber \\
    &\qquad \qquad \qquad \times \,
    \bigg(\, \log p(\by_t \!\mid\! \btheta, \bxi_t) \,-\, \log p(\by_t\! \mid\! \bxi_t) 
    \bigg)\,
    \mathrm{d}\by_{t} \mathrm{d}\bx_{0:t} \mathrm{d}\btheta \nonumber \\
    &\quad +\, \int \int \int p(\btheta\!\mid\! h_{t-1})   p(\bx_{0:t-1}\!\mid \!\btheta, h_{t-1}) f(\bx_t \!\mid\! \bx_{t-1}, \btheta, \bxi_t) \Big(\, \nabla_{\bxi_t} g(\by_t \!\mid\! \bx_t, \btheta, \bxi_t)\Big) \,\times \nonumber \\
    &\qquad \qquad \qquad \times \,
    \bigg( \,
    \log p(\by_t \!\mid\! \btheta, \bxi_t)\, - \,\log p(\by_t\! \mid\! \bxi_t) 
    \bigg)\,
    \mathrm{d}\by_{t} \mathrm{d}\bx_{0:t} \mathrm{d}\btheta \nonumber \\
    &\quad + \,\int \int \int p(\btheta\!\mid\! h_{t-1})   p(\bx_{0:t-1}\!\mid \!\btheta, h_{t-1}) f(\bx_t \!\mid\! \bx_{t-1}, \btheta, \bxi_t) g(\by_t \!\mid\! \bx_t, \btheta, \bxi_t) \,\times \nonumber \\
    &\qquad \qquad \qquad\times \,
    \bigg( \,
    \nabla_{\bxi_t}\! \log p(\by_t \!\mid\! \btheta, \bxi_t) \,-\, \nabla_{\bxi_t}\! \log p(\by_t\! \mid\! \bxi_t) 
    \bigg) \,
    \mathrm{d}\by_{t} \mathrm{d}\bx_{0:t} \mathrm{d}\btheta \nonumber.
\end{align}

Using the relation $\nabla p = p\,\nabla \log p$, we express each derivative of $f$ and $g$ in terms of their log-gradients.
This reformulation allows us to reassemble the entire expression as a single expectation over the joint distribution \(\Gamma(\by_t,\bx_{0:t},\btheta \!\mid \!\bxi_t, h_{t-1})\):
\begin{align}
    \nabla_{\bxi_t}\!\mathcal{I}(\bxi_t) 
    &= 
    \int \int \int p(\btheta\!\mid\! h_{t-1})   p(\bx_{0:t-1}\!\mid \!\btheta, h_{t-1})  f(\bx_t \!\mid\! \bx_{t-1}, \btheta, \bxi_t) g(\by_t \!\mid\! \bx_t, \btheta, \bxi_t)\, \times \nonumber \\
    &\qquad \qquad \times\, \nabla_{\bxi_t}\! \log f(\bx_t \!\mid\! \bx_{t-1}, \btheta, \bxi_t) \, \bigg(\,
    \log p(\by_t \!\mid\! \btheta, \bxi_t) \,-\, \log p(\by_t\! \mid\! \bxi_t) 
    \bigg)\,
    \mathrm{d}\by_{t} \mathrm{d}\bx_{0:t} \mathrm{d}\btheta \nonumber \\
    &\quad + \int \int \int p(\btheta\!\mid\! h_{t-1})   p(\bx_{0:t-1}\!\mid \!\btheta, h_{t-1}) f(\bx_t \!\mid\! \bx_{t-1}, \btheta, \bxi_t)  g(\by_t \!\mid\! \bx_t, \btheta, \bxi_t) \,\times \nonumber \\
    &\qquad \qquad \times\, \nabla_{\bxi_t}\! \log g(\by_t \!\mid\! \bx_t, \btheta, \bxi_t) \,
    \bigg(\,
    \log p(\by_t \!\mid\! \btheta, \bxi_t) \,-\, \log p(\by_t\! \mid\! \bxi_t) 
    \bigg) \,
    \mathrm{d}\by_{t} \mathrm{d}\bx_{0:t} \mathrm{d}\btheta \nonumber \\
    &\quad +\, \int \int \int p(\btheta\!\mid\! h_{t-1})   p(\bx_{0:t-1}\!\mid \!\btheta, h_{t-1}) f(\bx_t \!\mid\! \bx_{t-1}, \btheta, \bxi_t) g(\by_t \!\mid\! \bx_t, \btheta, \bxi_t) \,\times \nonumber \\
    &\qquad \qquad \times \,
    \bigg(\,
    \nabla_{\bxi_t}\! \log p(\by_t \!\mid\! \btheta, \bxi_t) \,-\, \nabla_{\bxi_t}\! \log p(\by_t\! \mid\! \bxi_t) 
    \bigg)\,
    \mathrm{d}\by_{t} \mathrm{d}\bx_{0:t} \mathrm{d}\btheta \nonumber\\
    &= \mathbb{E}_{\Gamma(\cdot \mid \bxi_t, h_{t-1})} 
    \bigg[ \,
    \nabla_{\bxi_t}\! \log f(\bx_t \!\mid\! \bx_{t-1}, \btheta, \bxi_t) \,
    \Big(\,
    \log p(\by_t \!\mid\! \btheta, \bxi_t) \,-\, \log p(\by_t\! \mid\! \bxi_t) 
    \Big)
    \nonumber \\
    &\qquad \qquad \qquad \qquad+ \,\nabla_{\bxi_t}\! \log g(\by_t \!\mid\! \bx_t, \btheta, \bxi_t)\, \Big(\, 
    \log p(\by_t \!\mid\! \btheta, \bxi_t) \,-\, \log p(\by_t\! \mid\! \bxi_t) 
    \Big) 
    \nonumber \\
    &\qquad  \qquad \qquad\qquad + \,
    \Big(\,
    \nabla_{\bxi_t}\! \log p(\by_t \!\mid\! \btheta, \bxi_t) \,-\, \nabla_{\bxi_t}\! \log p(\by_t\! \mid\! \bxi_t) 
    \Big)
    \bigg]
    \nonumber
\end{align}

Finally, using \(\nabla \log L = \frac{\nabla L}{L}\) and \(\nabla \log Z = \frac{\nabla Z}{Z}\), 
we obtain the compact gradient form reported in \eqref{eq:gradient_eig} (Section~\ref{subsec:eig-grad}):
\begin{align}
    \nabla_{\bxi_t}\mathcal{I}(\bxi_t) 
    &= \mathbb{E}_{\Gamma(\cdot \mid \bxi_t, h_{t-1})} 
    \Bigg[\,
        \frac{\nabla_{\bxi_t}\! L_{\btheta, \bxi_t}(\by_t)}{L_{\btheta,\bxi_t}(\by_t)} 
        \,-\, 
        \frac{\nabla_{\bxi_t}\! Z_{\bxi_t}(\by_t)}{Z_{\bxi_t}(\by_t)} 
        \nonumber \\
        & \qquad \qquad \qquad \qquad + \log \frac{L_{\btheta,\bxi_t}(\by_t)}{Z_{\bxi_t}(\by_t)}\, \Big(\, \nabla_{\bxi_t}\! \log f(\bx_t\! \mid \!\bx_{t-1}, \btheta, \bxi_t) \,+\, \nabla_{\bxi_t}\! \log g(\by_t\! \mid\! \bx_t, \btheta, \bxi_t) \Big)  
    \Bigg]. \nonumber
\end{align}

The quantities \(L_{\btheta,\bxi_t}(\by_t)\) and \(Z_{\bxi_t}(\by_t)\) correspond to the likelihood and evidence terms, respectively:
\begin{align}
    L_{\btheta,\bxi_t}(\by_t) &= \mathbb{E}_{p(\bx_{0:t}\mid \btheta, \bxi_t, h_{t-1})}
    \Big[\, g(\by_t\!\mid\!\bx_t,\btheta,\bxi_t)\Big], \nonumber
    \\[2mm] 
    Z_{\bxi_t}(\by_t) &= \mathbb{E}_{p(\bx_{0:t}, \btheta\mid \bxi_t, h_{t-1})} \Big[\, g(\by_t\!\mid\!\bx_t,\btheta,\bxi_t)\Big],
    \nonumber
\end{align}
where
\begin{align}
p(\bx_{0:t},\btheta\!\mid\!\bxi_t,h_{t-1})
&= f(\bx_t\!\mid\!\bx_{t-1},\btheta,\bxi_t)\,p(\bx_{0:t-1}\!\mid\!\btheta,h_{t-1}) \,p(\btheta\!\mid\!h_{t-1}),\nonumber
\\
p(\bx_{0:t}\!\mid\!\btheta,h_{t-1})
&= f(\bx_t\!\mid\!\bx_{t-1},\btheta,\bxi_t)\,p(\bx_{0:t-1}\!\mid\!\btheta,h_{t-1}), \nonumber
\end{align}
denote the joint and conditional predictive distributions at time~\(t\!-\!1\).

Following the same steps used above (product rule and log-gradients), the likelihood and evidence gradients are
\begin{align}
\nabla_{\bxi_t} L_{\btheta,\bxi_t}(\by_t)
&= \mathbb{E}_{p(\bx_{0:t}\mid \btheta,\bxi_t,h_{t-1})}
\Big[\,
\nabla_{\bxi_t} g(\by_t \!\mid\! \bx_t,\btheta,\bxi_t)
\,+\,
g(\by_t \!\mid\! \bx_t,\btheta,\bxi_t)\,
\nabla_{\bxi_t}\!\log f(\bx_t \!\mid\! \bx_{t-1},\btheta,\bxi_t)
\Big], \nonumber 
\\
\nabla_{\bxi_t} Z_{\bxi_t}(\by_t)
&= \mathbb{E}_{p(\bx_{0:t},\btheta\mid \bxi_t,h_{t-1})}
\Big[\,
\nabla_{\bxi_t} g(\by_t \!\mid\! \bx_t,\btheta,\bxi_t)
\,+\,
g(\by_t \!\mid\! \bx_t,\btheta,\bxi_t)\,
\nabla_{\bxi_t}\!\log f(\bx_t \!\mid\! \bx_{t-1},\btheta,\bxi_t)
\Big]. \nonumber
\end{align}

\section{Monte Carlo Approximations of Gradients}
\label{ap:approximations}

We now derive the Monte Carlo approximations used to estimate the gradient
\(\nabla_{\bxi_t}\mathcal{I}(\bxi_t)\)
in \eqref{eq:gradient_eig} (Section~\ref{subsec:eig-grad}).
The procedure mirrors the sampling strategy introduced in
Section~\ref{subsec:estimators} for the \gls*{eig} itself.

\paragraph{Outer expectation.}
The gradient expression in \eqref{eq:gradient_eig} (Section~\ref{subsec:eig-grad})
involves an expectation over the joint distribution
\(\Gamma(\by_t,\bx_{0:t},\btheta\mid\bxi_t,h_{t-1})\),
which can be approximated with \(L\) Monte Carlo samples:
\begin{align}
\widehat{\nabla_{\bxi_t}\mathcal{I}}(\bxi_t)
&=
\sum_{\ell=1}^{L} w_{\by,t}^{(\ell)}
\Bigg[\,\,
\frac{\widehat{\nabla_{\bxi_t} L}^N_{\btheta^{(\ell)},\bxi_t}(\widetilde{\by}_t^{(\ell)})}
     {\widehat{L}^N_{\btheta^{(\ell)},\bxi_t}(\widetilde{\by}_t^{(\ell)})}
\,-\,
\frac{\widehat{\nabla_{\bxi_t} Z}^{M,N}_{\bxi_t}(\widetilde{\by}_t^{(\ell)})}
     {\widehat{Z}^{M,N}_{\bxi_t}(\widetilde{\by}_t^{(\ell)})}
     \nonumber \\
&\qquad \qquad + \,
\log
\frac{\widehat{L}^{N}_{\btheta^{(\ell)},\bxi_t}(\widetilde{\by}_t^{(\ell)})}
     {\widehat{Z}^{M,N}_{\bxi_t}(\widetilde{\by}_t^{(\ell)})}
     \,\,
\Big(\,
\nabla_{\bxi_t}\log f(\widetilde{\bx}_t^{(\ell)} \!\mid\! {\bx}_{t-1}^{(\ell)}, \btheta_{t-1}^{(\ell)}, \bxi_t)
\,+ \,
\nabla_{\bxi_t}\log g(\widetilde{\by}_t^{(\ell)} \!\mid\! \widetilde{\bx}_t^{(\ell)}, \btheta_{t-1}^{(\ell)}, \bxi_t)\,
\Big)\,
\Bigg]. \label{eq:eig-grad-approx}
\end{align}

\paragraph{Sample generation.}
Monte Carlo samples from \(\Gamma(\cdot \mid \bxi_t, h_{t-1})\)
form the array
\[
\big\{(\btheta_{t-1}^{(m)},\,\bx_{0:t-1}^{(m,n)},\,\widetilde{\bx}_t^{(m,n)},\,\widetilde{\by}_t^{(m,n)})\big\},
\qquad
m=1,\dots,M,\;\; n=1,\dots,N,
\]
with weights combining the parameter and state particle weights from time~\(t\!-\!1\):
\[
w_{\by,t}^{(m,n)} = w_{\btheta,t-1}^{(m)} \, w_{\bx,t-1}^{(m,n)}.
\]

Each pair \((m,n)\) corresponds to a nested particle configuration:
\begin{align}
\btheta_{t-1}^{(m)} &\sim p(\btheta\!\mid\! h_{t-1}), \nonumber \\
\bx_{0:t-1}^{(m,n)} &\sim p(\bx_{0:t-1}\!\mid\! \btheta_{t-1}^{(m)},h_{t-1}), \nonumber \\
\widetilde{\bx}_t^{(m,n)} &\sim f(\cdot\!\mid\!\bx_{t-1}^{(m,n)},\btheta_{t-1}^{(m)},\bxi_t), \nonumber \\
\widetilde{\by}_t^{(m,n)} &\sim g(\cdot\!\mid\!\widetilde{\bx}_t^{(m,n)},\btheta_{t-1}^{(m)},\bxi_t). \nonumber
\end{align}

For notational simplicity, we relabel the samples and weights as
\begin{equation}\label{eq:samples_ell}
\big\{(\btheta_{t-1}^{(\ell)},\,\bx_{0:t-1}^{(\ell)},\,\widetilde{\bx}_t^{(\ell)},\,\widetilde{\by}_t^{(\ell)},\,w_{\by,t}^{(\ell)})\big\},
\qquad \ell = 1,\dots,L, \quad L = M \times N,
\end{equation}
where each index \(\ell\) corresponds to a unique pair \((m,n)\).
This reindexing allows for the compact summation form used in the estimator above.

\paragraph{Likelihood and evidence estimates.}
Each term in \eqref{eq:eig-grad-approx} (Appendix~\ref{ap:approximations})
requires evaluating—and differentiating—the likelihood and evidence.
Following the same nested sampling structure used for the \gls*{eig} estimators
(Section~\ref{subsec:estimators}),
we approximate these quantities using
\(M\) outer (parameter) samples and \(N\) inner (state) samples:
\begin{align}
\widehat{L}_{\btheta^{(\cdot)},\bxi_t}^{N}(\by_t)
&= 
\sum_{n=1}^{N} 
w_{\bx,t}^{(\cdot,n)}\,
g(\by_t\! \mid\! \ddot{\bx}_t^{(\cdot,n)}, \btheta^{(\cdot)}, \bxi_t) \nonumber
\\
&\qquad \qquad \text{for} \quad
\ddot{\bx}_t^{(\cdot,n)}\! \sim\! 
f(\cdot \!\mid\! \bx_{t-1}^{(\cdot,n)}, \btheta^{(\cdot)}, \bxi_t), \nonumber
\\[2mm]
\widehat{Z}_{\bxi_t}^{M,N}(\by_t)
&=
\sum_{m=1}^{M} \sum_{n=1}^{N} 
w_{\btheta,t}^{(m)}\, w_{\bx,t}^{(m,n)}\,
g(\by_t\! \mid\! \dot{\bx}_t^{(m,n)}, \dot{\btheta}^{(m)}_{t-1}, \bxi_t) \nonumber
\\
&\qquad \qquad \text{for} \quad
\dot{\btheta}^{(m)}_{t-1}\! \sim\! \kappa_M(\cdot \!\mid\! \btheta_{t-1}^{(m)}) ~~\text{and}~~
\dot{\bx}_t^{(m,n)} \!\sim\! 
f(\cdot \mid \bx_{t-1}^{(m,n)}, \dot{\btheta}^{(m)}, \bxi_t).\nonumber
\end{align}

Here, the superscripts \((n)\) and \((m,n)\) emphasize that the estimators depend
on the number of inner (state) and outer (parameter) particles, respectively.
The weighting terms \(w_{\bx,t}^{(m,n)}\) and \(w_{\btheta,t}^{(m)}\) correspond
to the normalized particle weights from the posterior approximations of the NPF at time~\(t-1\).

\paragraph{Gradients of likelihood and evidence.}
Following Appendix~\ref{ap:gradient},
the gradients of the inner expectations can be estimated as
\begin{align}
    \widehat{\nabla_{\bxi_t} L}_{\btheta^{(\cdot)},\bxi_t}^N(\by_t) &\,=\, 
     \sum_{n=1}^N w_{\bx,t-1}^{(\cdot,n)} \,\Big[ \,
    \nabla_{\bxi_t} g({\by}_t \!\mid \!\ddot{\bx}_t^{(\cdot,n)}, \btheta, \bxi_t)
   \,+\, g({\by}_t \!\mid\! \ddot{\bx}_t^{(\cdot,n)}, \btheta,\bxi_t) \, \nabla_{\bxi_t} \!\log f(\ddot{\bx}_t^{(\cdot,n)}\! \mid\! {\bx}_{t-1}^{(\cdot,n)}, \btheta, \bxi_t) 
    \,\Big], \nonumber
    \\
    \widehat{\nabla_{\bxi_t} Z}_{\bxi_t}^{M,N}(\by_t)
&\,=\, 
    \sum_{m=1}^M \sum_{n=1}^N w_{\bx,t-1}^{(m,n)} \,w_{\btheta,t-1}^{(m)} \,\Big[ \,
    \nabla_{\bxi_t} g({\by}_t \!\mid\! \dot{\bx}_t^{(m,n)}, \dot{\btheta}_{t-1}^i, \bxi_t) \nonumber
    \\
    & \qquad \qquad \qquad
   \,+\, g({\by}_t \!\mid\! \dot{\bx}_t^{(m,n)}, \dot{\btheta}_{t-1}^i, \bxi_t) \, \nabla_{\bxi_t}\! \log f(\dot{\bx}_t^{(m,n)} \!\mid\! {\bx}_{t-1}^{(m,n)}, \dot{\btheta}_{t-1}^i, \bxi_t) 
    \,\Big]. \nonumber
\end{align}

The first expression provides a Monte Carlo estimate of the gradient of the likelihood \(p(\by_t\!\mid\!\btheta,\bxi_t)\) under fixed parameter particles,
while the second corresponds to the evidence gradient averaged over both parameter and state particles. Together, they yield estimates of the terms required in the gradient of the \gls*{eig}.


\section{Proof of Consistency of the EIG Estimator}
\label{ap:consistency}

Let $D_\theta \subset \mathbb{R}^{d_\theta}$ denote the compact parameter domain.
For a bounded measurable function $h : D_\theta \to \mathbb{R}$ we write $h \in {B}(D_\theta)$, with supremum norm 
$\|h\|_\infty = \sup_{\btheta \in D_\theta} |h(\btheta)|$.
For a distribution $\phi$ on a measurable space $\mathcal{X}$ and a test function $f : \mathcal{X} \to \mathbb{R}$, we denote $(f,\phi) = \int f(x)\,\phi(\mathrm{d}x)$.
We use $\|\cdot\|_p$ for the $\ell^p$ norm on $\mathbb{R}^{d_\theta}$, and abbreviate almost sure convergence as $\xrightarrow{\text{a.s.}}$.

\begin{theorem}[Consistency of {EIG} estimator]
Let $\widehat{\mathcal{I}}(\bxi_t)$ denote the \gls*{nmc} estimator of the expected information gain defined in \eqref{eq:eig_estimator} (Section~\ref{subsec:estimators}), constructed with $M$ parameter particles, $N$ state particles per parameter, and $L$ pseudo-observations. Under Assumptions~\ref{as:kernel scaling}–\ref{as:nmc},
\[
\widehat{\mathcal{I}}(\bxi_t) \;\xrightarrow[L,M,N \to \infty]{\text{a.s.}}\; \mathcal{I}(\bxi_t),
\]
for any $t$ and $\bxi_t$.
\end{theorem}

\begin{assumption}[Jittering kernel scaling]\label{as:kernel scaling}
There exist $p \ge 1$ and $c_\kappa < \infty$ such that
\[
\sup_{\btheta' \in D_\theta} \int \|\btheta - \btheta'\|_p \, \kappa_M(d\btheta \!\mid\! \btheta')
\;\le\; \frac{c_\kappa^p}{M^{p/2}}.
\]
\end{assumption}

\begin{assumption}[Jittering kernel regularity]\label{as:kernel regularity}
For any $h \in {B}(D_\theta)$,
\[
\sup_{\btheta' \in D_\theta} \int |h(\btheta) - h(\btheta')| \, \kappa_M(d\btheta \!\mid\! \btheta')
\;\le\; \frac{c_\kappa \|h\|_\infty}{\sqrt{M}}.
\] 
\end{assumption}

\begin{assumption}[Lipschitz dependence of state posteriors on $\btheta$]\label{as:lipschitz}
For each $t \ge 1$, let $\phi_{t,\btheta}$ denote the posterior of $\bx_t$ given $(\by_{1:t},\bxi_{1:t})$ and parameter $\btheta \in D_\theta$.
Then for every bounded measurable $f : \mathbb{R}^{d_x} \to \mathbb{R}$ there exists $b_t < \infty$ such that
\[
\big| (f,\phi_{t,\btheta'}) - (f,\phi_{t,\btheta''}) \big|
\;\le\; b_t \, \|f\|_\infty \, \|\btheta' - \btheta''\|,
\quad \forall\, \btheta',\btheta'' \in D_\theta.
\]
\end{assumption}

\begin{assumption}[Bounded, positive likelihood]\label{as:likelihood}
For (almost) every $\by_t$ in the observation space and each $t$,
\[
0 < \inf_{\btheta \in D_\theta,\, \bx_t} g(\by_t \!\mid\! \bx_t, \btheta, \bxi_t) 
\;\le\; \sup_{\btheta \in D_\theta,\, \bx_t} g(\by_t \!\mid\! \bx_t, \btheta, \bxi_t) < \infty.
\]
That is, $g$ is uniformly bounded above and bounded away from zero across $\btheta \in D_\theta$ (and $\bx_t$) for any $\bxi_t$.
\end{assumption}

\begin{assumption}[Regularity of the integrand for NMC]\label{as:nmc}
Let $L_{\btheta,\bxi_t}(\by_t) = \mathbb{E}_{p(\bx_{0:t}\mid \btheta,h_{t-1})}[\,g(\by_y \!\mid \!\bx_t,\btheta,\bxi_t)\,]$ and 
$Z_{\bxi_t}(\by_t)=\mathbb{E}_{p(\btheta\mid h_{t-1})p(\bx_{0:t}\mid \btheta,h_{t-1})}[\,g(\by_t\! \mid\! \bx_t,\btheta,\bxi_t)\,]$.
The information–gain integrand 
\[
f(\by_t,\btheta,\bx_{0:t}) \;=\; \log\!\frac{L_{\btheta,\bxi_t}(\by_t)}{Z_{\bxi_t}(\by_t)}
\]
is (i) Lipschitz continuous in the inner random arguments passed from the inner estimators (i.e., in the latent state path and any inner re-sampled $\btheta$) and (ii) square-integrable ($f \in L^2$).
Assumption~\ref{as:likelihood} implies (ii), since boundedness and positivity yield $L_{\btheta,\bxi_t},Z_{\bxi_t} \in (\epsilon,\infty)$ and hence $\log(L_{\btheta,\bxi_t}/Z_{\bxi_t})\in L^2$ and is continuous on $[\epsilon,\infty)$.
\end{assumption}

\medskip
\begin{remark}
Assumptions~\ref{as:kernel scaling}–\ref{as:likelihood} are standard in the analysis of NPF 
\citep{Crisan18bernoulli} 
and ensure that the empirical measures of the particle system converge almost surely to the true posterior/predictive distrubutions.
Assumption~\ref{as:nmc} is the usual \gls*{nmc} regularity condition 
\citep{Rainforth18}
, stated here for the \gls*{eig} integrand.
\end{remark}

\subsection{Proof of Theorem C.1}

\paragraph{Convergence of the evidence estimator.}
Theorem~3 of 
\citet{Crisan18bernoulli} 
shows that the joint parameter--state posterior empirical measure produced by the NPF,
\(\widehat{\pi}_t^{M,N} = \widehat{p}(\mathrm{d}\btheta, \mathrm{d}\bx_{0:t}\!\mid\! h_t)\),
converges to the true posterior
\(\pi_t = p(\mathrm{d}\btheta, \mathrm{d}\bx_{0:t}\!\mid\! h_t)\) as \(M,N\to\infty\).
Adapting notation to our setting:

\begin{theorem}[Crisan and Míguez, 2018, Thm.~3]
\label{th:npf_convergence}
Let \(h_T=\{\by_{1:T},\bxi_{1:T}\}\) be fixed and \(f\in B(D_\theta\times\mathbb{R}^{d_x})\).
Under Assumptions~\ref{as:kernel scaling}--\ref{as:likelihood}, for any \(p\ge 1\) and \(1\le t\le T\),
\[
\big\| (f,\widehat{\pi}_t^{M,N}) - (f,\pi_t) \big\|_p
~\le~
\frac{c_t \|f\|_\infty}{\sqrt{M}} + \frac{\bar{c}_t \|f\|_\infty}{\sqrt{N}},
\]
where the constants \(\{c_t,\bar{c}_t\}_{1\le t\le T}\) are finite and independent of \(M,N\).
\end{theorem}

Taking \(f(\btheta,\bx_{0:t})=g(\by_t\!\mid\! \bx_t,\btheta,\bxi_t)\), we obtain
\begin{equation}\label{eq:Z-consistency}
\widehat{Z}^{M,N}_{\bxi_t}(\by_t)
~=~ (f,\widehat{\pi}_t^{M,N})
~\xrightarrow[M,N\to\infty]{\text{a.s.}}~
(f,\pi_t)
~=~ Z_{\bxi_t}(\by_t),
\end{equation}
so the particle estimator of the evidence converges almost surely to the true value. In particular, any Monte Carlo estimator that integrates against \(\widehat{\pi}_t^{M,N}\) (including its gradient forms) inherits these asymptotic guarantees.

\paragraph{Convergence of the likelihood estimator.}
Theorem~2 and Remark~10 of 
\citet{Crisan18bernoulli} 
establish convergence for the empirical parameter posterior \(\widehat{\mu}_t^{M,N}\!=\!\widehat{p}(\mathrm{d}\btheta\!\mid\! h_t)\), and for the conditional state filters \(\widehat{\phi}^N_{t,\btheta'}=\widehat{p}(\mathrm{d}\bx_t\!\mid \!\btheta', h_t)\) computed within the NPF.

\begin{theorem}[Crisan and Míguez, 2018, Thm.~2]
\label{th:npf_param_posterior}
Let \(h_T=\{\by_{1:T},\bxi_{1:T}\}\) be fixed (\(T<\infty\)) and \(h\in B(\mathbb{R}^{d_\theta})\).
Under Assumptions~\ref{as:kernel scaling}--\ref{as:likelihood}, for any \(p\ge 1\) and \(1\le t\le T\),
\[
\big\| (h,\widehat{\mu}_t^{M,N}) - (h,\mu_t) \big\|_p
~\le~
\frac{c_t \|h\|_\infty}{\sqrt{M}} + \frac{\bar{c}_t \|h\|_\infty}{\sqrt{N}},
\]
where \(\widehat{\mu}_t^{M,N}=\widehat{p}(\mathrm{d}\btheta\!\mid\! h_t)\) and
\(\mu_t=p(\mathrm{d}\btheta\!\mid\! h_t)\), and the constants \(\{c_t,\bar{c}_t\}_{1\le t\le T}\) are finite and independent of \(M,N\).
\end{theorem}

Moreover, by Remark~10 in 
\citet{Crisan18bernoulli}
, the same proof yields uniform error bounds for the conditional state filters \(\widehat{\phi}^N_{t,\btheta'}\) associated with each parameter particle: letting \(\phi_{t,\btheta'}=p(\mathrm{d}\bx_t\!\mid\! \btheta',h_t)\) and any \(f\in B(\mathbb{R}^{d_x})\),
\[
\sup_{1\le m\le M}
\big\| (f,\widehat{\phi}^N_{t,\btheta'}) - (f,\phi_{t,\btheta'}) \big\|_p
~\le~
\frac{k_t \|f\|_\infty}{\sqrt{M}} + \frac{\bar{k}_t \|f\|_\infty}{\sqrt{N}},
\]
for some finite constants \(k_t,\bar{k}_t\) independent of \(M,N\).
Choosing \(f(\bx_t)=g(\by_t\!\mid\! \bx_t,\btheta,\bxi_t)\), we obtain, for each \(m\) parameter sample in the NPF,
\begin{equation}\label{eq:L-consistency}
\widehat{L}^{N}_{\btheta^{(m)},\bxi_t}(\by_t)
~=~ (f,\widehat{\phi}^N_{t,\btheta^{(m)}})
~\xrightarrow[N\to\infty]{\text{a.s.}}~
(f,\phi_{t,\btheta^{(m)}})
~=~ L_{\btheta^{(m)},\bxi_t}(\by_t).
\end{equation}

\paragraph{Convergence of the \gls*{nmc} estimator.}
Finally, Theorem~1 of 
\citet{Rainforth18} 
gives consistency of \gls*{nmc}
estimators under mild regularity. In our setting, the outer level averages over
\(\Gamma(\by_t,\bx_{0:t},\btheta\!\mid \!\bxi_t,h_{t-1})\) and the inner levels compute
\(L_{\btheta,\bxi_t}(\by_t)\) and \(Z_{\bxi_t}(\by_t)\). Up to some rewriting, the theorem is:

\begin{theorem}[Rainforth et al., 2018, Thm.~1]
\label{th:nmc_rainforth}
Let \(f\big(y,\gamma_L(y),\gamma_Z(y)\big)=\log\big(\gamma_L(y)/\gamma_Z(y)\big)\),
with \(\gamma_L(y)=\mathbb{E}_{p(x)}[g(y\!\mid \!x)]\) and
\(\gamma_Z(y)=\mathbb{E}_{p(x,\theta)}[g(y\!\mid\! x,\theta)]\).
If \(f\) is Lipschitz and \(f\big(\,y,\gamma_L(y),\gamma_Z(y)\,\big),\,\,g(y\!\mid\! x),\,\,g(y\!\mid\! x,\theta)\in L^2\), then the \gls*{nmc} estimator
\[
\widehat{I}^{L,M,N}
= \frac{1}{L}\sum_{\ell=1}^L
f\!\left(
y^{(\ell)},
\frac{1}{N}\sum_{n=1}^N g(y^{(\ell)}\!\mid\! x^{(\ell,n)}),
\frac{1}{MN}\sum_{m=1}^M\sum_{n=1}^N g(y^{(\ell)}\!\mid\! x^{(\ell,m,n)},\theta^{(\ell,m)})
\right)
~\xrightarrow[L,M,N\to\infty]{\text{a.s.}}~ I
\]
\end{theorem}

In our formulation, \(g\) is the observation model
\(g(\by_t\mid \bx_t,\btheta,\bxi_t)\), which uses latent states \(\bx_t\) for the likelihood term, and uses both \((\bx_t,\btheta)\) for the evidence.
Assumption~\ref{as:nmc} (regularity of the integrand) ensures Lipschitz continuity and square-integrability.
Combining \eqref{eq:Z-consistency}–\eqref{eq:L-consistency} (Appendix~\ref{ap:consistency}) with Theorem~\ref{th:nmc_rainforth}, we conclude that, for any \(t\) and \(\bxi_t\),
\[
\widehat{\mathcal{I}}(\bxi_t)
~\xrightarrow[L,M,N\to\infty]{\text{a.s.}}~
\mathcal{I}(\bxi_t).
\]
\qed

\section{Algorithm Details}
\label{ap:algorithm}

This appendix provides complete details of BAD-PODS and the baselines we compare with.
Algorithm~\ref{alg:badpods_full} presents the full BAD-PODS pipeline, including stochastic gradient–based design optimization and sequential inference via an NPF.
At each time $t$, BAD-PODS optimizes the design by maximizing the \gls*{eig} using stochastic gradient ascent (SGA),
executes the experiment at that design, and updates the joint parameter–state posterior.


\begin{algorithm}[tb]
\caption{BAD-PODS: Bayesian adaptive design for partially observable dynamical systems}
\label{alg:badpods_full}
\begin{algorithmic}[1]
\REQUIRE Particle counts $(M,N)$; inner \gls*{sga} steps $K$ with stepsizes $\{\eta_k\}_{k=0}^{K-1}$; priors $p(\btheta),\,p(\bx_0)$
\STATE \textbf{Initialization:}
$\btheta_0^{(m)}\!\sim\!p(\btheta)$, $\,\bx_0^{(m,n)}\!\sim\!p(\bx_0)$;
$\,w_{\btheta,0}^{(m)}\!=\!1/M$, $\,w_{\bx,0}^{(m,n)}\!=\!1/N$

\FOR{$t=1$ to $T$}
  \STATE Initialize $\bxi_t^{(0)}$ randomly \algcmt{design optimization loop}
  \FOR{$k=0$ to $K-1$}
    \STATE Draw outer MC samples $(\widetilde{\by}_t^{(\ell)},\widetilde{\bx}_t^{(\ell)},\btheta_{t-1}^{(\ell)})\! \sim\! \Gamma(\cdot\!\mid\!\bxi_t^{(k)},h_{t-1})$ ({Appendix~\ref{ap:approximations}})
    \STATE For each $m$, sample $\ddot{\bx}_t^{(m,n)} \sim f(\cdot\mid \bx_{t-1}^{(m,n)}, \btheta_{t-1}^{(m)}, \bxi_t^{(k)})$ for $n=1{:}N$ \algcmt{for $\widehat L$}
    \STATE Jitter $\dot{\btheta}_{t-1}^{(m)} \sim \kappa_M(\cdot\mid \btheta_{t-1}^{(m)})$ and sample $\dot{\bx}_t^{(m,n)} \sim f(\cdot\mid \bx_{t-1}^{(m,n)}, \dot{\btheta}_{t-1}^{(m)}, \bxi_t^{(k)})$ \algcmt{for $\widehat Z$}
    \STATE Compute $\widehat{L}_{\btheta_{t-1},\,\bxi_t^{(k)}}^N$ and $\widehat{Z}_{\bxi_t^{(k)}}^{M,N}$ in \eqref{eq:likelihood_estimator}--\eqref{eq:evidence_estimator} in Section~\ref{subsec:estimators}
    \STATE Compute $\widehat{\nabla_{\bxi_t}\mathcal{I}}(\bxi_t^{(k)})$ in \eqref{eq:eig-grad-approx} in Appendix~\ref{ap:approximations}
    \STATE $\bxi_t^{(k+1)} \gets \bxi_t^{(k)} + \eta_k\, \widehat{\nabla_{\bxi_t}\mathcal{I}}(\bxi_t^{(k)})$
  \ENDFOR

  \STATE $\bxi_t^\star \gets \bxi_t^{(K)}$; collect $\by_t \sim g(\cdot\mid \bxi_t^\star)$ \algcmt{execute design, collect data}
  \STATE Jitter parameters: $\btheta_{t|t-1}^{(m)} \sim \kappa_M(\cdot\mid \btheta_{t-1}^{(m)})$ 
  \FOR{$m=1$ to $M$}
  \STATE Propagate states: $\bx_{t|t-1}^{(m,n)} \sim f(\cdot\mid \bx_{t-1}^{(m,n)}, \btheta_{t|t-1}^{(m)}, \bxi_t^\star)$
  \STATE Update state weights and normalize them
  \[
  w_{\bx,t}^{(m,n)} \propto w_{\bx,t-1}^{(m,n)}\, g\!\left(\by_t \mid \bx_{t|t-1}^{(m,n)}, \btheta_{t|t-1}^{(m)}, \bxi_t^\star\right), \qquad \qquad \widetilde{w}_{\bx,t}^{(m,n)} = \frac{w_{\bx,t}^{(m,n)}}{\sum_{j=1}^N w_{\bx,t}^{(m,j)}}
  \]
  \STATE Resample states: Draw indices $n_1,\ldots,n_N$ with probability $\widetilde{w}_{\bx,t}^{(m,1)},\ldots,\widetilde{w}_{\bx,t}^{(m,N)}$
  \\ Set $\bx_t^{(m,n)}=\bx_{t|t-1}^{(m,n_j)}$, for $j=1:N$ \algcmt{update conditional state posterior (NPF)}
  \ENDFOR
  \STATE Update parameter weights and normalize them:
  \[
  w_{\btheta,t}^{(m)} = w_{\btheta,t-1}^{(m)} \sum_{n=1}^N w_{\bx,t}^{(m,n)}, \qquad \qquad \widetilde{w}_{\btheta,t}^{(m)} = \frac{w_{\btheta,t}^{(m)}}{\sum_{i=1}^M w_{\btheta,t}^{(i)}}
  \]
  \STATE Resample parameters: Draw indices $m_1,\ldots,m_M$ with probability $\widetilde{w}_{\btheta,t}^{(1)},\ldots,\widetilde{w}_{\btheta,t}^{(M)}$
  \\ Set $\btheta_t^{(m)}=\btheta_{t|t-1}^{(m_i)}$, for $i=1:M$ \algcmt{update parameter posterior (NPF)}
\ENDFOR
\end{algorithmic}
\end{algorithm}

For comparison, the baselines share the same inference update as BAD-PODS and differ only in design selection:
\begin{itemize}
  \item \textbf{Random design:} identical to Algorithm~\ref{alg:badpods_full} but without the optimization loop. Each $\bxi_t^\star$ is sampled randomly.
  \item \textbf{Oracle (grid search):} identical to Algorithm~\ref{alg:badpods_full} but replaces the stochastic-gradient loop with an exhaustive search over a fixed grid of candidate designs. At each time step, we evaluate the estimated \gls*{eig} for all grid points and set $\bxi_t^\star$ to the maximiser. This provides an expensive but informative upper-bound reference in low-dimensional design spaces.
  \item \textbf{Static BED:} performs a single offline optimization of the whole sequence $\{\bxi_1,\dots,\bxi_T\}$ and then runs the same online inference using those fixed designs.
\end{itemize}

\section{Two-group SIR Model}
\label{ap:sir}

\subsection{Model Description}

While the formulation applies to an arbitrary number of subpopulations, we focus here on the
two-group case. Let the stacked state be
\(\mathbf{X}(\tau) = \big(S^{(1)}(\tau),\, I^{(1)}(\tau),\, S^{(2)}(\tau),\, I^{(2)}(\tau)\big)^\top \in \mathbb{R}^4\), where $S$ and $I$ stand for \emph{susceptible} and \emph{infectious}, respectively.
Populations \(N_g\) are constant, so the number of \emph{recovered} is \(R^{(g)}(\tau) = N^{(g)} - S^{(g)}(\tau) - I^{(g)}(\tau)\).
Parameters are \(\btheta = \{(\beta^{(g)},\gamma^{(g)})\}_{g=1}^2\) are infection and recovery rate constants for each group \(g\). Cross‐group mixing is specified by a fixed
\(\mathbf{M} \in \mathbb{R}^{2\times 2}\) (e.g., nonnegative, rows summing to one).

Given the current state \(\mathbf{X}\), the corresponding total (population-level) transition rates are
\begin{equation}
  \lambda^{(g)}(\mathbf{X})
  \;=\;
  \beta^{(g)}\,S^{(g)} \sum_{h=1}^2 \mathbf{M}_{g h}\,\frac{I^{(h)}}{N^{(h)}}
  \qquad \text{and} \qquad
  r^{(g)}(\mathbf{X})
  \;=\;
  \gamma^{(g)}\,I^{(g)} .
  \label{eq:rates-2g}
\end{equation}
where \(\lambda^{(g)}(\mathbf{X})\) is the total rate of new infections and
\(r^{(g)}(\mathbf{X})\) the total recovery rate in group \(g\).

\paragraph{\Gls*{sde}.}
Let \(\mathbf{W}(\tau)\in\mathbb{R}^4\) be a standard Wiener process. The Itô \gls*{sde} is
\begin{equation}
  \mathrm{d}\mathbf{X}(\tau)
  \;=\;
  f\big(\mathbf{X}(\tau)\big)\,\mathrm{d}\tau
  \;+\;
  G\big(\mathbf{X}(\tau)\big)\,\mathrm{d}\mathbf{W}(\tau),
  \label{eq:sir-sde-2g-app}
\end{equation}
with drift \(f(\mathbf{X}) = S\,a(\mathbf{X})\) and diffusion factor
\(G(\mathbf{X}) = S\,\mathrm{diag}\!\big(\sqrt{a(\mathbf{X})}\big)\), where
\[
  S \;=\;
  \begin{pmatrix}
   -1 &  0 &  0 &  0 \\
    1 & -1 &  0 &  0 \\
    0 &  0 & -1 &  0 \\
    0 &  0 &  1 & -1
  \end{pmatrix}
  \qquad \text{and} \qquad
  a(\mathbf{X}) \;=\;
  \bigg(\lambda^{(1)}(\mathbf{X}),\, r^{(1)}(\mathbf{X}),\, \lambda^{(2)}(\mathbf{X}),\, r^{(2)}(\mathbf{X})\bigg)^\top .
\]

\paragraph{Euler–Maruyama discretization.}
For step \(\Delta\tau>0\),
\begin{equation}
  \mathbf{X}_{t+1}
  \;=\;
  \mathbf{X}_{t}
  \;+\;
  S\,a(\mathbf{X}_{t})\,\Delta\tau
  \;+\;
  S\,\mathrm{diag}\!\big(\sqrt{a(\mathbf{X}_{t})}\big)\,\Delta\mathbf{W}_{t},
  \qquad
  \Delta\mathbf{W}_{t} \sim \mathcal{N}\!\big(\mathbf{0},\,\Delta\tau\,\mathbf{I}_4\big).
  \label{eq:em-2g-app}
\end{equation}
To preserve feasibility, project to the per‐group simplex of counts:
\begin{align*}
  &S_{t+1}^{(g)} \leftarrow \min\!\big\{N^{(g)},\,\max\{0,\,S_{t+1}^{(g)}\}\big\}, \\
  &I_{t+1}^{(g)} \leftarrow \min\!\big\{N^{(g)} - S_{t+1}^{(g)},\,\max\{0,\,I_{t+1}^{(g)}\}\big\},
\end{align*}
for $g\in\{1,2\}$, and optionally set \(R_{t+1}^{(g)} = N^{(g)} - S_{t+1}^{(g)} - I_{t+1}^{(g)}\).

\paragraph{Observation model and design.}
At time \(t\), choose \(\bxi_t=\big(\xi_t^{(1)},\xi_t^{(2)}\big)\) with \(\xi_t^{(g)}\ge 0\) and
\(\xi_t^{(1)}+\xi_t^{(2)}=1\), splitting a fixed sampling effort \(\kappa>0\).
We observe incident counts with Poisson noise:
\begin{equation}
  y_t^{(g)} \,\big|\, \mathbf{X}_t, \btheta, \bxi_t
  \;\sim\;
  \mathrm{Poisson}\!\left(\lambda^{\text{obs}}_{t,(g)}\right),
  \qquad
  \lambda^{\text{obs}}_{t,(g)}
  \;=\;
  \kappa\,\xi_t^{(g)}\,\rho^{(g)}\,\frac{I_t^{(g)}}{N^{(g)}},
  \quad g\in\{1,2\},
  \label{eq:obs-2g-app}
\end{equation}
where \(\rho^{(g)}>0\) is a known detection scale. 

Intuitively, the design vector $\bxi_t$ specifies how the fixed sampling effort
$\kappa$ is distributed across groups, controlling the expected number of observations in each subpopulation. While it does not affect the underlying epidemic dynamics, it modulates the observation noise level and therefore the information gained about the unknown parameters.

\subsection{Simulation Setup}

We consider two subpopulations of equal size, \( N^{(1)} \!=\! N^{(2)}\! =\! 200\), each initialized with \(I_0^{(1)} \!=\! I_0^{(2)} \!=\! 5\) infectious individuals and \(S_0^{(g)} = N^{(g)} - I_0^{(g)}\). The per-group detection scales are \(\rho= \big(\rho^{(1)},\rho^{(2)}\big) = (0.95, 0.5)\), and the total sampling effort is fixed to \(\kappa \!=\! 100\).
The cross-group mixing matrix is
\[
\mathbf{M} =
\begin{pmatrix}
0.9 & 0.1 \\
0.1 & 0.9
\end{pmatrix},
\]
which induces moderate within-group interaction and limited cross-group coupling.
The continuous-time dynamics are integrated with an Euler–Maruyama step
\(\Delta\tau\! =\! 0.1\).

The true epidemiological parameters are \(\beta^{(1)}\! =\! 0.65,\, \gamma^{(1)} \!=\! 0.15\) (for group~1, treated as unknown) and
\(\beta^{(2)} \!=\! 0.55,\, \gamma^{(2)}\! =\! 0.15\) (for group~2, fixed and known).
We infer only \(\btheta = \big(\beta^{(1)}, \gamma^{(1)}\big)\) with uniform priors
\[
\beta^{(1)}, \gamma^{(1)} \sim \mathcal{U}(0.1, 1.0).
\]

\vspace{1mm}


\paragraph{Algorithmic settings.}
For the NPF, we use \(M\!=\!N\!=\!100\) particles for parameters and states, respectively (\(L\!=\!M\!\times\! N\) total Monte Carlo samples).
We constrained the particle counts to \(M\!=\!N\) and selected this value via grid search over \(\{50,100,200,300,400,500\}\), choosing the smallest particle count for which performance gains became marginal relative to the added computational cost.
Parameter jittering is
$\kappa_M(\cdot\!\mid\!\btheta')=\mathcal{N}(\btheta\!\mid\!\btheta',\sigma_{\text{jitter}}^2 \boldsymbol{I}_{d_\theta})$, with variance scaled as
\[
\sigma_{\text{jitter}}^2 = \frac{c_{\text{jitter}}}{M^{3/2}}.
\]
The scale constant $c_{\text{jitter}}$ was tuned by a two-stage grid search: first a coarse grid \(\{0.1,0.5,1,5,10\}\), followed by a refined grid \(\{1,2,3,4,5\}\) around the best-performing region. We set $c_{\text{jitter}}\!=\!2$.
We resample parameter particles at every time step in all experiments, although using an \gls*{ess} threshold to trigger resampling is also compatible with this algorithm.

In the BAD-PODS implementation, design optimisation at each step uses stochastic gradient ascent with \(K\!=\!500\) updates, where \(K\) was selected via grid search over \(\{50,100,200,300,400,500,600\}\).
We use ADAM \citep{DBLP:journals/corr/KingmaB14} with standard momentum parameters \(\beta_1\!=\!0.9\), \(\beta_2\!=\!0.999\), and \(\varepsilon\!=\!10^{-6}\), and we tune the step size by grid search over \(\alpha\in\{5\times10^{-4},10^{-3},2.5\times10^{-3},5\times10^{-3},10^{-2},2\times10^{-2},3\times10^{-2},4\times10^{-2},5\times10^{-2},7.5\times10^{-2}\}\), selecting \(\alpha\!=\!0.03\).
Design variables are initialised uniformly,
\(\xi_t^{(1)}\sim\mathcal{U}(0,1)\) and \(\xi_t^{(2)}=1-\xi_t^{(1)}\), and mapped through a sigmoid reparameterisation to enforce simplex constraints. For the oracle implementation, instead of stochastic optimisation we do a grid search. We discretise $\xi_t^{(1)}\in[0,1]$ into 500 candidate values, to obtain a fine grid resolution and select that design that maximises the \gls*{eig}.

All experiments are repeated for \(T\!=\!200\) sequential design steps and averaged over 50 Monte Carlo realizations with independent random seeds.

\section{Moving Source Location Model}
\label{ap:source}

\subsection{Model Description}

\paragraph{State and parameters.}
The latent state is \(\bx_t=(p_{x,t},p_{y,t},\phi_t)^\top \in \mathbb{R}^2 \times (-\pi,\pi]\), where \(\bp_t=(p_{x,t},p_{y,t})^\top\) is position in a plane and \(\phi_t\) heading angle. Static motion parameters are \(\btheta=(v_x,v_y,v_\phi)\).

\paragraph{Dynamics (transition).}
With sampling step \(\Delta t>0\),
\begin{equation}
\bx_t \;=\; \bx_{t-1}
\;+\; \Delta t \begin{bmatrix}
v_x \cos \phi_{t-1}\\[2pt]
v_y \sin \phi_{t-1}\\[2pt]
v_\phi
\end{bmatrix}
\;+\; \bepsilon_t,
\qquad
\bepsilon_t \sim \mathcal{N}\!\big(\mathbf{0},\,\mathbf{Q}\big),
\label{eq:app-source-dyn}
\end{equation}
with \(\mathbf{Q}=\mathrm{diag}(\sigma_x^2,\sigma_y^2,\sigma_\phi^2)\). After propagation the heading is wrapped to the principal interval, \(\phi_t \leftarrow \mathrm{wrap}_\pi(\phi_t)\), where \(\mathrm{wrap}_\pi(\phi)=((\phi+\pi)\bmod 2\pi)-\pi\). The transition \eqref{eq:app-source-dyn} does not depend on the design.

\paragraph{Sensors and design.}
There are \(J\) fixed sensors at positions \(\{\bs_j\}_{j=1}^J\subset\mathbb{R}^2\).
At time \(t\) the design is the vector of orientations \(\bxi_t=(\xi_{t,1},\ldots,\xi_{t,J})\in[-\pi,\pi)^J\).
We define the bearing from sensor \(j\) to the source as \(\psi_{t,j}(\bp_t)=\operatorname{atan2}\!\big((\bp_t-\bs_j)_y,(\bp_t-\bs_j)_x\big)\), and the angular mismatch \(\Delta_{t,j}=\xi_{t,j}-\psi_{t,j}(\bp_t)\).

\paragraph{Observation model.}
Each sensor reports a log–intensity corrupted by \gls*{iid}\ Gaussian noise,
\begin{align}
\log y_{t,j} \,\big|\, \bp_t,\btheta,\bxi_t &\sim
\mathcal{N}\!\big(\log \mu_{t,j},\,\sigma^2\big), \label{eq:app-source-like}\\
\mu_{t,j} &= b \;+\; \frac{\alpha_j}{\,m + \|\bp_t-\bs_j\|^2\,}\, D\!\big(\Delta_{t,j}\big),
\label{eq:app-source-mu}
\end{align}
for $j=1,\ldots,J$, where \(b>0\) is a background level, \(m>0\) a saturation constant,
\(\alpha_j\ge 0\) a per-sensor strength, and
\[
D(\delta)\;=\;\bigg(\frac{1+d\cos\delta}{1+d}\bigg)^k, \qquad d\in[0,1), \quad k>1,
\]
is a cardioid directivity function that favors alignment between the sensor orientation and the bearing to the source, while ensuring that \(\mu_{t,j} > 0\) for all configurations. The design vector \(\bxi_t\) enters the observation model \eqref{eq:app-source-mu} only through the angular offset \(\Delta_{t,j}\).

\subsection{Simulation Setup}

We use \(J=2\) fixed sensors at \(\bs_1\!=\!(3,0)^\top\) and \(\bs_2\!=\!(0,3)^\top\).
Per–sensor strengths are \(\alpha_j\!=\!5\) (for \(j\!=\!1,2\)), with background level \(b\!=\!0.1\) and saturation constant \(m\!=\!0.1\).
The angular directivity $D(\delta)$ is set with \(d\!=\!1\) and \(k\!=\!4\).
Observations are log–intensities with \gls*{iid} Gaussian noise \(\sigma^2\!=\!0.1\), hence \(R\!=\!\sigma^2 \mathbf{I}_J\).
Designs are initialized uniformly as \(\bxi_t\!\sim\!\mathcal{U}([-\pi,\pi))^J\).

For the state transition we use step \(\Delta t\!=\!0.1\) and process noise
\(\bepsilon_t\!\sim\!\mathcal{N}(\mathbf{0},Q)\) with \(Q\!=\!\mathrm{diag}(0.2,\,0.2,\,10^{-2})\).
We infer only the planar velocity components \(\btheta=(v_x,v_y)\), with uniform priors
\(v_x,v_y\sim\mathcal{U}(0.5,1.5)\).
The true (data–generating) parameters are \(v_x^\star\!=\!v_y^\star\!=\!1.0\) (with \(v_\phi\!=\!0.3\) known).


\paragraph{Algorithmic settings.}
We use an NPF with \(M\!=\!N\!=\!300\) particles for parameters and states (total \(L\!=\!M\!\times\! N\) Monte Carlo samples per design step).
We constrained the particle counts to \(M\!=\!N\) and selected this value via grid search over \(\{50,100,200,300,400,500\}\), choosing the smallest particle count for which performance gains became marginal relative to the added computational cost.
Parameter jittering is
$\kappa_M(\cdot\!\mid\!\btheta')=\mathcal{N}(\btheta\!\mid\!\btheta',\sigma_{\text{jitter}}^2 \boldsymbol{I}_{d_\theta})$, with
\[
\sigma_{\text{jitter}}^2 = \frac{c_{\text{jitter}}}{M^{3/2}}.
\]
The scale constant \(c_{\text{jitter}}\) was tuned by a two-stage grid search: a coarse grid \(\{0.1,0.5,1,5,10\}\), followed by a refined grid \(\{0.05,0.1,0.15,0.2,0.25\}\) around the best-performing region; we set \(c_{\text{jitter}}\!=\!0.15\).
We resample parameter particles at every time step.

In the BAD-PODS implementation, design optimisation at each step uses stochastic gradient ascent with \(K=300\) updates, where \(K\) was selected via grid search over \(\{50,100,200,300,400,500,600\}\).
We use ADAM \citep{DBLP:journals/corr/KingmaB14} with \(\beta_1\!=\!0.9\), \(\beta_2\!=\!0.999\), and \(\varepsilon\!=\!10^{-6}\), and we tune the step size over \(\alpha\in\{5\times10^{-4},10^{-3},2.5\times10^{-3},5\times10^{-3},10^{-2},2\times10^{-2}\}\), selecting \(\alpha\!=\!0.01\).
For the oracle implementation, we discretise the 2D design space $(-\pi, \pi] \!\times\! (-\pi, \pi]$ into a $25\!\times\! 25$ grid (625 candidate values), selecting the one that maximises the \gls*{eig}.

All experiments are repeated for \(T\!=\!50\) sequential design steps and averaged over 50 Monte Carlo realizations with independent random seeds.

\section{Ecological Growth Model}
\label{ap:growth}

\subsection{Model Description}

We consider a stochastic logistic growth model with harvesting, commonly used in ecological population dynamics \citep{zhou2009modified}. The latent state is the (scaled) population size $x_t \in \mathbb{R}_+$ at discrete time $t$.
The design $\xi_t \in [0,1]$ controls harvesting intensity (e.g., fishing effort or trapping rate), and therefore affects both the transition dynamics and the observation model.

\paragraph{Dynamics (transition).}
Let $\Delta>0$ be the integration step size. Conditioned on $(x_t,\xi_t)$ and parameters
$\btheta=(r,k)$, the dynamics are
\begin{equation}
x_{t+1}
=
x_t
+
\Delta\Bigl(
r\,x_t\bigl(1-\tfrac{x_t}{k}\bigr)
- q\,\xi_t\,x_t
\Bigr)
+
\varepsilon_t,
\qquad
\varepsilon_t \sim \mathcal{N}(0,\Delta\sigma_x^2),
\label{eq:growth-transition-app}
\end{equation}
where $r>0$ is the intrinsic growth rate, $k>0$ is the carrying capacity, and $q>0$ is the catchability coefficient. The term $q\,\xi_t\,x_t$ is the (expected) harvested amount per unit time, proportional to both effort $\xi_t$ and available population $x_t$.

\paragraph{Observation model and design.}
We define the harvested amount
\begin{equation}
\lambda_t = q\,\xi_t\,x_t.
\label{eq:growth-lambda-app}
\end{equation}
Observations provide a noisy, partially saturated measurement of this harvested amount:
\begin{equation}
y_t \!\mid\! x_t,\xi_t,\btheta
\sim
\mathcal{N}\!\bigl(h(\lambda_t),\,\sigma_y^2\bigr),
\qquad
h(\lambda_t)
=
C_{\max}\,\frac{\lambda_t}{C_{\text{half}}+\lambda_t},
\label{eq:growth-observation-app}
\end{equation}
where $C_{\max}>0$ is the saturation level and $C_{\text{half}}>0$ controls the transition from a linear to a saturated regime. This nonlinear response implies that increasing harvesting effort does not increase information indefinitely: for large $\lambda_t$ the function $h(\lambda_t)$ saturates, obtaining virtually same outcomes for different harvesting efforts.

The design is one-dimensional, $\xi_t\in[0,1]$. Small $\xi_t$ yields weak signals $\lambda_t$ and therefore low signal-to-noise, while large $\xi_t$ can quickly depress the latent state $x_t$ through the transition \eqref{eq:growth-transition-app} and also pushes $h(\lambda_t)$ into a saturation regime. Together, these effects create a non-trivial trade-off and can yield intermediate optimal designs.

\subsection{Simulation Setup}
\label{ap:growth-setup}

We simulate the growth/harvest model in \eqref{eq:harvest-transition}--\eqref{eq:harvest-observation} over a horizon of \(T\!=\!20\) steps, with integration step \(\Delta\!=\!0.1\). The initial state is set to \(x_0 \!=\! 0.4\,k\). The true (data-generating) parameters are \(r^\star \!=\! 0.5\) and \(k^\star \!=\! 300\). We assume process noise variance \(\sigma_x^2 \!=\! 0.1\), observation noise variance \(\sigma_y^2 \!=\! 2.0\), and saturation parameters \(C_{\max}\!=\!90\) and \(C_{\text{half}}\!=\!30\) are known. The design is the harvest effort \(\xi_t \in [0,1]\), initialised at each time step as \(\xi_t \sim \mathcal{U}(0,1)\).
We infer only \(\btheta\!=\!(r,k)\), with independent priors \(r\sim \mathcal{U}(0.2,1.2)\) and \(k\sim \mathcal{U}(200,800)\).

\paragraph{Algorithmic settings.}
We use an NPF with \(M\!=\!N\!=\!200\) particles for parameters and states (total \(L\!=\!M\!\times\! N\) Monte Carlo samples per design step).
We constrained the particle counts to \(M\!=\!N\) and selected this value via grid search over \(\{50,100,200,300,400,500\}\), choosing the smallest particle count for which performance improvements became marginal relative to the added computational cost.
Parameter jittering is
\(\kappa_M(\cdot\!\mid\!\btheta')\!=\!\mathcal{N}\!\big(\btheta\!\mid\!\btheta',\Sigma_{\text{jitter}} \big)\), with
\[
\Sigma_{\text{jitter}} = \begin{bmatrix}
    \sigma_{{r}}^2 & 0 \\
    0 & \sigma_{{k}}^2
\end{bmatrix}, \qquad
\sigma_{{r}}^2 = \frac{c_{{r}}}{M^{3/2}}, \qquad
\sigma_{{k}}^2 = \frac{c_{{k}}}{M^{3/2}}.
\]
The scale constants \(c_r\) and \(c_k\) were tuned by a two-stage grid search: a coarse grid $\{0.1,0.5,1,5,10\}$ and $\{1,5,10,20,40,80\}$, followed by a refined grid around the best-performing region. We set $c_r\!=\!0.05$ and $c_k\!=\!50$. We resample parameter particles at every time step.

In the BAD-PODS implementation, design optimisation at each step uses stochastic gradient ascent with \(K\!=\!200\) updates, where \(K\) was selected via grid search over \(\{50,100,200,300,400,500,600\}\).
We use ADAM \citep{DBLP:journals/corr/KingmaB14} with \(\beta_1\!=\!0.9\), \(\beta_2\!=\!0.999\), and \(\varepsilon\!=\!10^{-6}\), and we tune the step size over \(\alpha\in\{5\times10^{-4},10^{-3},2.5\times10^{-3},5\times10^{-3},10^{-2},2\times10^{-2},3\times10^{-2},4\times10^{-2},5\times10^{-2}\}\), selecting \(\alpha\!=\!5\times10^{-3}\).
For the oracle baseline, we discretise the 1D design space \([0,1]\) into a grid of 500 candidates and select \(\xi_t\) by maximising the estimated \gls*{eig} over this grid at each time step.

All experiments are averaged over 50 Monte Carlo realisations with independent random seeds.

\section{Computational Infrastructure}
\label{ap:infrastructure}

All experiments were conducted on a high-performance computing (HPC) cluster equipped with NVIDIA V100 GPUs.
The computations made use of multi-core Intel Xeon processors, high-capacity DDR4 memory, and InfiniBand interconnects for fast data transfer between nodes.
Each GPU node contained multiple V100 devices (16–32 GB memory per GPU), and all optimization and inference workloads were executed using these resources.




\end{document}

%% file: acronyms.tex
\usepackage[acronym]{glossaries}
\newacronym{new}{BAD-PODS}{Bayesian adaptive design for partially observable dynamical systems}
\newacronym{bca}{BCa}{bias-corrected and accelerated}
\newacronym{abc}{ABC}{approximate Bayesian computation}
\newacronym{AWGN}{AWGN}{addtive white {G}aussian noise}
\newacronym{apf}{APF}{auxiliary particle filter}

\newacronym{bad}{BAD}{Bayesian adaptive design}
\newacronym{bed}{BED}{Bayesian experimental design}
\newacronym{boed}{BOED}{Bayesian optimal experimental design}

\newacronym{ckf}{CKF}{cubature Kalman filter}

\newacronym{dpf}{DPF}{differentiable particle filter}

\newacronym{eig}{EIG}{expected information gain}
\newacronym{ekf}{EKF}{extended Kalman filter}
\newacronym{enkf}{EnKF}{ensemble Kalman filter}
\newacronym{ess}{ESS}{effective sample size}

\newacronym{bim}{BIM}{Bayesian information matrix}

\newacronym{hmm}{HMM}{heterogeneous multi-scale model}

\newacronym{iid}{i.i.d.}{independent and identically distributed}
\newacronym{ipf}{IPF}{island particle filter}

\newacronym{kf}{KF}{Kalman filter}
\newacronym{kl}{KL}{Kullback-Leibler}

\newacronym{mi}{MI}{mutual information}
\newacronym{mse}{MSE}{mean square error}

\newacronym{nhf}{NHF}{nested hybrid filter}
\newacronym{ngf}{NGF}{nested Gaussian filter}
\newacronym{nmc}{NMC}{nested Monte Carlo}
\newacronym{npf}{NPF}{nested particle filter}
\newacronym{nsmc}{NSMC}{nested sequential Monte Carlo}
\newacronym{nmse}{NMSE}{normalized mean square error}
\newacronym{pce}{PCE}{prior contrastive estimation}
\newacronym{pdf}{pdf}{probability density function}
\newacronym{pmcmc}{PMCMC}{particle Markov chain Monte Carlo}
\newacronym{pf}{PF}{particle filter}
\newacronym{pl}{PL}{particle learning}
\newacronym{pde}{PDE}{partial differential equation}

\newacronym{qmc}{QMC}{quasi Monte Carlo}
\newacronym{qkf}{QKF}{quadrature Kalman filter}

\newacronym{rv}{r.v.}{random variable}
\newacronym{rml}{RML}{recursive maximum likelihood}

\newacronym{sg}{SG}{stochastic gradient}
\newacronym{sga}{SGA}{stochastic gradient ascent}
\newacronym{sir}{SIR}{susceptible-infectious-recovered}
\newacronym{smc}{SMC}{sequential Monte Carlo}
\newacronym{smc2}{SMC$^2$}{sequential Monte Carlo squared}
\newacronym{sqmc}{SQMC}{sequential quasi-Monte Carlo}
\newacronym{sde}{SDE}{stochastic differential equation}
\newacronym{stpf}{ST-PF}{space-time particle filter}
\newacronym{ssm}{SSM}{state-space model}
\newacronym{ukf}{UKF}{unscented Kalman filter}
\newacronym{ut}{UT}{unscented transform}

\newacronym{vnmc}{VNMC}{variational nested Monte Carlo}

\newacronym{wrt}{w.r.t.}{with respect to}


%% file: macros.tex
\newcommand{\bp}{\boldsymbol{p}}

\newcommand{\bs}{\boldsymbol{s}}

\newcommand{\bx}{\boldsymbol{x}}
\newcommand{\by}{\boldsymbol{y}}

\newcommand{\btheta}{\boldsymbol{\theta}}

\newcommand{\bxi}{\boldsymbol{\xi}}

\newcommand{\bepsilon}{{\boldsymbol{\epsilon}}}

\definecolor{forestgreen}{RGB}{22, 187, 110}
\definecolor{darkred}{RGB}{242, 81, 27}

\newcommand{\black}{\textcolor{black}}

\newcommand{\reals}{\mathbb R}      
\newcommand{\naturals}{\mathbb N}

\newcommand{\algcmt}[1]{\hfill$\triangleright$~\textit{#1}}